\documentclass{article}

\PassOptionsToPackage{numbers, compress}{natbib}

\usepackage[preprint]{neurips_2026}

\usepackage[utf8]{inputenc} %
\usepackage[T1]{fontenc}    %
\usepackage{wrapfig}
\usepackage{url}            %
\usepackage{booktabs}       %
\usepackage{amsfonts}       %
\usepackage{nicefrac}       %
\usepackage{microtype}      %
\usepackage{xcolor}
\usepackage[table]{xcolor}
\usepackage{booktabs}
\usepackage{etoc}
\usepackage{titletoc}
\usepackage[dvipsnames]{xcolor}
\usepackage{subcaption}
\usepackage{suffix}

\usepackage{amsmath,amssymb,stmaryrd,mathtools,amsthm}
\usepackage{xcolor}
\usepackage{xspace}
\usepackage{bbm}
\usepackage{graphicx}
\usepackage{lipsum}

\definecolor{perfblue}{RGB}{64, 114, 175}
\usepackage[colorlinks=true,allcolors=perfblue]{hyperref}

\ifdefined\pdfminorversion
\fi
\definecolor{dark_red}{RGB}{122, 0, 0}
\definecolor{coral}{RGB}{255, 119, 94}
\definecolor{pink_orange}{RGB}{255, 72, 126}
\definecolor{vibrant_pink}{RGB}{255, 0, 104}
\definecolor{pink_pink}{RGB}{255, 37, 153}
\definecolor{wine}{RGB}{204, 0, 102}

\definecolor{light_orange}{RGB}{255, 198, 107}
\definecolor{orange(sae/ece)}{rgb}{1.0, 0.49, 0.0}
\definecolor{dark_orange}{RGB}{216,92,0}

\definecolor{org-purp-0}{RGB}{165, 76, 0}
\definecolor{org-purp-1}{RGB}{250, 130, 28}
\definecolor{org-purp-2}{RGB}{226, 89, 68}
\definecolor{org-purp-3}{RGB}{206, 92, 124}
\definecolor{org-purp-4}{RGB}{116, 80, 146}
\definecolor{org-purp-5}{RGB}{110, 78, 157}

\definecolor{teal(sae/ece)}{rgb}{0, 0.47, 0.52}
\definecolor{aqua}{RGB}{52,172,139}
\definecolor{dark_aqua}{RGB}{35,115,93}
\definecolor{dark_green}{RGB}{0, 92, 34}

\definecolor{grape}{RGB}{112,48,160}
\definecolor{purple}{rgb}{0.74, 0.65, 1.0}
\definecolor{dark_purple}{rgb}{0.58, 0.0, 0.82}
\definecolor{periwinkle}{RGB}{191, 140, 230}

\definecolor{light_gray}{rgb}{0.9, 0.9, 0.9}
\definecolor{medium_gray}{rgb}{0.6, 0.6, 0.6} 
\definecolor{dark_gray}{rgb}{0.2, 0.2, 0.2} 

\definecolor{sky_blue}{RGB}{37, 166, 213}
\definecolor{light_blue}{rgb}{0.33, 0.80, 1}
\definecolor{dark_blue}{rgb}{0.098, 0.239, 0.52}
\definecolor{ocean}{RGB}{13, 121, 202}
\definecolor{light_ocean}{RGB}{18, 178, 235}
\definecolor{dark_ocean}{RGB}{10, 89, 148}
\definecolor{vibrant_blue}{RGB}{14, 120, 255}

\definecolor{dark_brown}{rgb}{0.3255, 0.004, 0.001}

\newcommand{\para}[1]{\noindent\textbf{#1.}}

\newcounter{qnum}
\newcounter{tnum}
\newcommand{\ours}{\textcolor{black}{\texttt{WEAVER}}\xspace}
\newcommand{\ctrlworld}{\textcolor{black}{Ctrl-World}\xspace}
\newcommand{\ctrlworldft}{\textcolor{black}{Ctrl-World-FT}\xspace}
\newcommand{\oursft}{\textcolor{black}{\texttt{WEAVER-FT}}\xspace}
\newcommand{\oursreflow}{\textcolor{black}{\texttt{WEAVER-REFLOW}}\xspace}

\newcommand{\realdataset}{\mathcal{D}_{\text{real}}}

\newcommand{\obs}{o}
\newcommand{\img}{I}
\newcommand{\proprio}{q}

\newcommand{\enc}{\mathcal{E}_\psi}
\newcommand{\dec}{\mathcal{D}_\eta}
\newcommand{\dyn}{f_\phi}

\newcommand{\lang}{\ell}

\newcommand{\latent}{z}
\newcommand{\mem}{\texttt{mem}}
\newcommand{\hist}{\texttt{hist}}
\newcommand{\latentmem}{\latenttraj^{\mem}}
\newcommand{\latenthist}{\latenttraj^{\hist}}
\newcommand{\latenttraj}{\mathbf{z}}
\newcommand{\latentpredtraj}{\mathbf{\hat{z}}}
\newcommand{\latentSpace}{\mathcal{Z}}

\newcommand{\policy}{\pi} %
\newcommand{\action}{a}
\newcommand{\acttraj}{\mathbf{\action}}

\newcommand{\actionSpace}{\mathcal{A}}

\newcommand{\reward}{R}
\newcommand{\critic}{V}

\definecolor{pastellavender}{HTML}{EDE7F6}
\definecolor{pastelmint}{HTML}{E0F2F1}
\definecolor{pastelblue}{HTML}{E3F2FD}
\definecolor{deeppurple}{HTML}{5E35B1}
\definecolor{tablegray}{HTML}{E5E7EB}
\definecolor{deepteal}{HTML}{00796B}
\definecolor{deepblue}{HTML}{1565C0}
\definecolor{lightpurpleborder}{HTML}{B39DDB}
\definecolor{lightteal}{HTML}{B2DFDB}
\definecolor{lightblueborder}{HTML}{90CAF9}
\definecolor{tableheader}{HTML}{E8EAF6}
\definecolor{lightblue}{rgb}{0.88, 0.95, 1.0}
\definecolor{slategray}{HTML}{2F4F4F}

\title{\raisebox{-0.15em}{\includegraphics[height=0.9em]{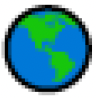}} \ours, Better, Faster, Longer: \\ An Effective World Model for Robotic Manipulation}

\author{%
    Arnav Kumar Jain\thanks{Equal Contribution. Correspondence to Arnav <arnav-kumar.jain@mila.quebec> and Yilin <yilinwu@andrew.cmu.edu>.} \\
  Mila - Qu\'ebec AI Institute\\
  Universit\'e de Montr\'eal \\
  \And
  Yilin Wu$^*$ \\
  Carnegie Mellon University\\
  \And
  Jesse Farebrother\\
  Mila - Qu\'ebec AI Institute\\
  McGill University \\
  \\
  \And
  Gokul Swamy \\
  Carnegie Mellon University\\
  \And
  Andrea Bajcsy \\
  Carnegie Mellon University\\
}

\begin{document}

\maketitle

\begin{abstract}
The potential impacts of world models (WMs, i.e., learned simulators) on robotics are far-reaching---policy evaluation, policy improvement, and test-time planning---all with limited real-world interaction. To unlock these downstream capabilities, a WM needs to jointly satisfy three desiderata: \textit{(i)} fidelity (i.e., producing simulated trajectories that correlate with reality), \textit{(ii)} consistency (i.e., producing simulated trajectories that are coherent over long horizons), and \textit{(iii)} efficiency (i.e., producing simulated trajectories quickly). We propose \ours (World Estimation Across Views for Embodied Reasoning): a WM architecture that simultaneously achieves all three desiderata, providing state-of-the-art results on robotic manipulation tasks. \ours is a multi-view WM trained to predict future latents and reward values via a flow-matching loss. 
We distill the key design decisions across model architecture, memory, and prediction objectives required to unlock the kinds of long-horizon dynamic manipulation tasks that have confounded prior world modeling approaches.
We apply \ours in robotic hardware, demonstrating its effectiveness at policy evaluation ($\rho=0.870$ correlation with real-world success rate), policy improvement (real-world success rate improvement of $38\%$ on top of the $\pi_{0.5}$ robot foundation model), and test-time planning (real-world success rate improvement of $14\%$ with a $5-10\times$ speedup over prior WMs). \ours also demonstrates better performance than prior WMs when evaluated on out-of-distribution scenarios. Code, models, and videos at: \href{https://arnavkj1995.github.io/WEAVER/}{\texttt{https://arnavkj1995.github.io/WEAVER/}}.

\end{abstract}

\section{Introduction}

World models (WMs, \citep{ha2018worldmodels}), or learned simulators, have attracted intense interest from both academia \cite{guo2026ctrlworld, zhou2026dino, yin2026playworld, quevedo2025worldgymworldmodelenvironment} and industry \cite{bruce2024genie, russell2503gaia}. This is because of the tremendous promise of WMs for robotics: the ability to both evaluate and improve policies without costly and often unsafe real-world interaction. Furthermore, WMs unlock test-time scaling when incorporated into planning algorithms.

To simultaneously deliver on the three promises of evaluation, improvement, and planning, a robot WM must jointly satisfy three core desiderata. The first is \textit{(i) fidelity}: producing physically accurate predictions that correlate with real-world outcomes. The second is \textit{(ii) consistency}: producing predictions that remain coherent over long prediction horizons. The third is \textit{(iii) efficiency}: producing predictions quickly. For example, policy evaluation and improvement require high-fidelity predictions (for handling arbitrary, visuomotor robot policies) as well as consistency (to handle multi-stage tasks). Relatedly, planning requires fast inference for dealing with the real-time requirements of robots.

Despite rapid progress, no existing robot WM satisfies all three desiderata in tandem. For example, video generation models \cite{mei2026video} produce high fidelity generations at the cost of low efficiency. Similarly, JEPA-style WMs \cite{assran2023self} have latent states that may not be decodable into the images required to evaluate arbitrary visuomotor robot policies. And while Dreamer-v4 \cite{hafner2025training} appears promising, learning an encoder from scratch rather than using a pretrained model can harm out-of-distribution robustness.

When we focus on robotic manipulation, the world modeling problem becomes even more complex, as we must handle multiple views of the scene, infer occluded objects from history, and ensure relatively high fidelity predicted world states rather than just visual aesthetics. 
Handling these complexities often comes at the cost of efficiency, with state-of-the-art WMs for manipulation like Ctrl-World \cite{guo2026ctrlworld} operating at far slower speeds than the real world, precluding their use in test-time planning and making policy improvement computationally challenging.

In response, we introduce  \ours (\textbf{W}orld \textbf{E}stimation \textbf{A}cross \textbf{V}iews for \textbf{E}mbodied \textbf{R}easoning): a WM architecture that achieves \textit{(i)} high fidelity, \textit{(ii)} long-horizon consistency, and \textit{(iii)} efficient generation, unlocking state-of-the-art performance across policy evaluation, improvement, and test-time planning on challenging robotic manipulation tasks. 
To achieve this trifecta of capabilities, \ours fuses together key design decisions from prior world modeling approaches. From the video generation community, we adopt diffusion forcing \cite{chen2024diffusion} and flow matching \cite{lipman2022flow} (for long-horizon generation at fast inference speeds) and the use of a pretrained encoder \cite{rombach2022high} (for out-of-distribution robustness). From latent world models \cite{hafner2025training, sharma2026world, guo2026ctrlworld}, we adopt the use of a reward prediction head to facilitate efficient evaluation without the need for an external judge model like a VLM. From JEPA \cite{assran2025v}, we adopt future latent prediction (rather than image reconstruction) as our primary training objective. Lastly, to handle the particular complexities of robot manipulation, we adopt the multi-view generation and memory architecture of Ctrl-World~\cite{guo2026ctrlworld}. 

Put together, we end up with a gestalt whole: a WM for robotic manipulation that can be used flexibly across evaluation, improvement, and planning. On a suite of five manipulation tasks (from pick and place to deformable object manipulation) performed on real hardware, \ours demonstrates strong correlation  ($\rho=0.870$) with real-world success rate when used for evaluation, improves the real-world success rate of the $\pi_{0.5}$ \cite{intelligence2025pi_} robot foundation model by $38\%$ \textit{without} any real-world interaction, and unlocks test-time planning $5-10\times$ faster than Ctrl-World~\cite{guo2026ctrlworld}.

\begin{figure}[t]
  \centering
  \includegraphics[width=\linewidth]{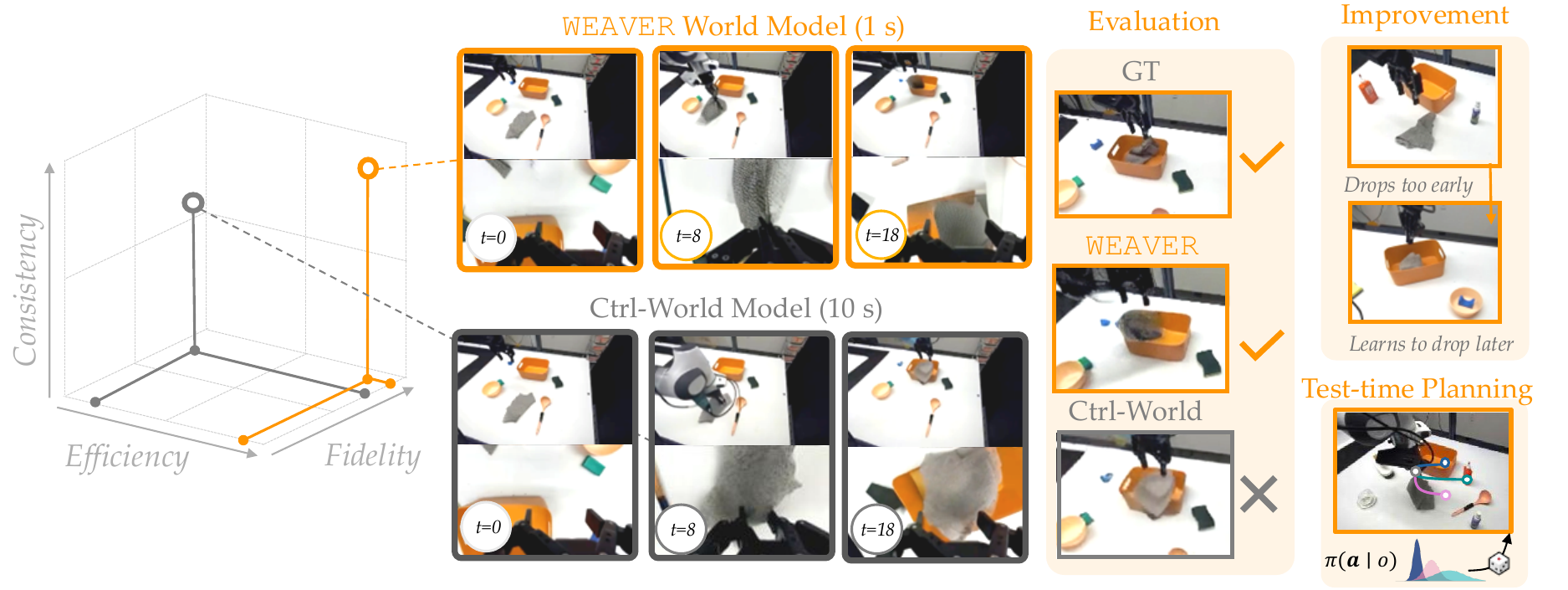}
  \caption{We present \ours, a \texttt{WM} that satisfies three desiderata: (i) high fidelity, (ii) long-horizon consistency and (iii) efficient generation. With these, we unlock the potential for downstream policy evaluation (middle), policy improvement (top right) and Test-time Planning (bottom right). }
\end{figure}

\section{Related Work}

\para{Robot World Models} 
While world models have been explored across autonomous driving~\cite{russell2503gaia, wang2024drivedreamer}, video games~\cite{he2025matrix}, and code generation~\cite{copet2025cwm},
we focus on their application to robotics ~\cite{wu2023daydreamer, guo2026ctrlworld, sharma2026world, team2025evaluating, assran2025v} -- more specifically visual manipulation. While improvements in video generation ~\cite{wiedemer2025video, blattmann2023stable} have lead to high \textit{(i)} fidelity WMs \cite{mei2026video, guo2026ctrlworld, sharma2026world, quevedo2025worldgymworldmodelenvironment, guo2026ctrlworld, gao2026dreamdojo, russell2503gaia}, these WMs are often not \textit{(iii)} efficient enough to use for test-time planning. However, incorporating key ingredients from the broader vision community, like flow matching \cite{lipman2022flow}, diffusion forcing \cite{chen2024diffusion} allows us to improve the \textit{(iii)} efficiency of \ours. Furthermore, the use of pretrained video generation model encoders \cite{rombach2022high} enhances \ours's robustness to out-of-distribution visual inputs, while the use of pretrained decoders allows us to evaluate arbitrary visuomotor robot policies unlike JEPA-style models \citep{assran2025v}. Finally, we adopt the latent reward and value heads of Dreamer-v4 ~\cite{hafner2025training} to enable \textit{(iii)} efficient evaluation and planning without the need to pass decoded images to an external and often slow VLM judge like in \cite{guo2026vlaw}.

Prior WMs~\cite{hafner2020Dream,hafner2021mastering,hansen2022tdmpc,wu2023daydreamer, wu2024forewarn, jain2026a, jain2022learning} struggle to maintain temporal \textit{(ii)} consistency across long horizons. In response, we adopt the use of multi-view prediction, history, and memory from ~\cite{guo2026ctrlworld, russell2503gaia} to ensure generations remain coherent even when gripper-object interactions are under occlusions. This is in contrast to earlier WMs like WorldGym~\cite{quevedo2025worldgymworldmodelenvironment}, DreamerV4~\cite{hafner2025training} and DreamDojo~\cite{gao2026dreamdojo}.

Perhaps the most similar approaches to our own are Ctrl-World \citep{guo2026ctrlworld} and Dreamer-v4 ~\cite{hafner2025training}. By using techniques from the video generation community \cite{chen2024diffusion,lipman2022flow} for more \textit{(iii)} efficient inference, we are able to produce higher \textit{(i)} fidelity generations that are more temporally \textit{(ii)} coherent in less time, Pareto dominating Ctrl-World \citep{guo2026ctrlworld}. By using a pretrained encoder \cite{rombach2022high} instead of learning one from scratch as in Dreamer-v4 ~\cite{hafner2025training}, we likely inherit better robustness to out-of-distribution visual inputs.

\para{Uses of World Models in Robotics} 
World models promise ``downstream'' advances in robotic policy evaluation, improvement, and test-time planning.
Prior work has shown that sufficiently faithful world models can enable scalable policy evaluation~\cite{team2025evaluating, yin2026playworld, wang2026interactive}, while early results suggest that synthetic trajectories may also improve policies \cite{guo2026vlaw, wang2026interactive}, though the extent to which this is true remains an open question. 
More recently, world models have been explored for test-time planning~\cite{qiplanning,wu2024forewarn}, where the central challenge is generating accurately quickly for online optimization. 
\ours is designed with each of these downstream applications in mind for robotic manipulation.

\section{\ours: 
World Estimation Across Views for Embodied Reasoning}
\begin{figure}
    \centering
    \includegraphics[width=\linewidth]{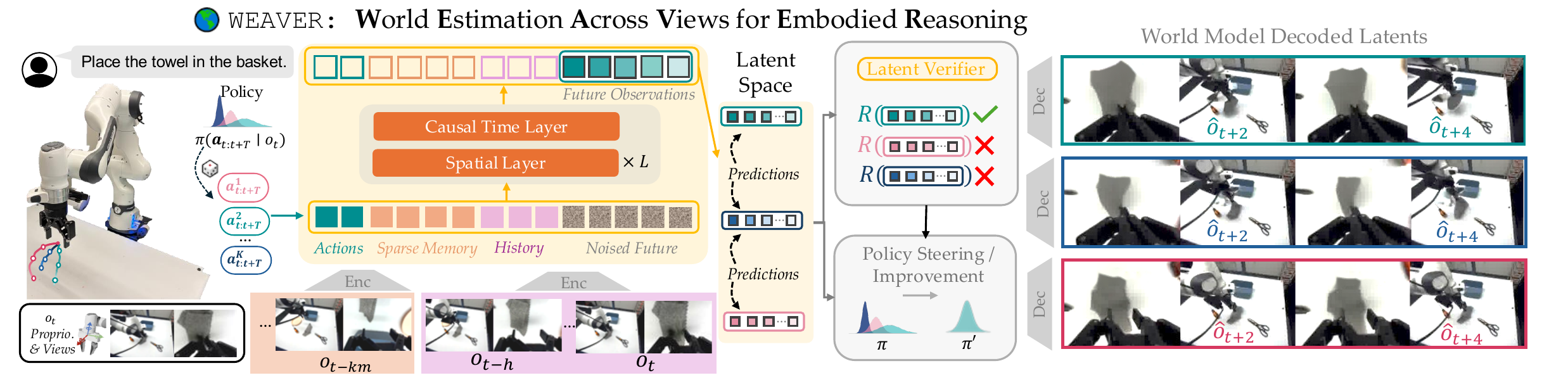}
    \caption{\textbf{\ours Architecture.} 
Left: The world model encodes memory, history, and action sequences to image future rollouts in latent space. 
Middle: The latent verifier, equipped with reward and critic heads, selects samples with high advantage to steer the policy distribution. 
Right: Decoded generation corresponding to different outcomes of action sequences.}
\label{fig:method_fig}
\end{figure}
We now describe the key ingredients in \ours: a robot world model designed to support policy evaluation, policy improvement, and test-time planning. These downstream applications of the WM on manipulation tasks imply three key desiderata upstream: \textit{(i) fidelity} across multiple views during physical interaction, \textit{(ii) consistent} predictions across long-horizon interactions that can introduce occlusions, and \textit{(iii) efficient} enough generation for use in a real-time planning algorithm.

To jointly satisfy these three desiderata, \ours fuses together a variety of ingredients. We first describe the key WM design decisions and training objective (Sec.~\ref{sec:training_objective}), followed by inference acceleration (Sec.~\ref{sec:faster_inference}) and latent-space value estimation (Sec.~\ref{sec:value_estimate}). We then show how, when put together, these components enable evaluation, improvement, and planning (Sec.~\ref{sec:downstream_application}).

\para{Setup: Robot \& Policy}
We consider long-horizon robotic manipulation tasks specified by a natural language instruction $\lang \in \mathcal{L}$. 
Let the robot's proprioceptive state (e.g., joint angles) be denoted by $\proprio \in \mathbb{R}^8$. The robot also has $n$ RGB views of the scene (e.g., from wrist and third person cameras); let this set of multi-view images be $\mathbf{\img} := (\img^1,\ldots,\img^n)$. 
At timestep $t$, the robot observes both the multiview images and proprioceptive state: $\obs_t := (\mathbf{\img}_t,\proprio_t) \in \mathcal{O}$.
Let the robot's action be denoted by $\action \in \actionSpace$ (e.g., joint velocities). 
Given any $(\obs_t,\lang)$, the robot's \textit{base policy} $\pi_{\theta}$ generates $\acttraj_t \sim \policy_\theta(\cdot \mid \obs_t,\lang)$, an $h$-step future action chunk (i.e., $\acttraj_t := \action_{t:t+h}$), which is then executed in the WM / environment.

\para{World Model Architecture} 
Our WM maps an observation 
$\obs_{t}$ into a latent state $\latent_t \in \latentSpace$ via a pretrained encoder $\latent_t \sim \enc(\obs_{t})$. 
A key design choice is conditioning our world model on both a \textit{memory} of every $k$th prior latent, $\latentmem_t := (\dots, \latent_{t-2k}, \latent_{t-k})$, as well as a $m$-step \textit{history} of the $m$ most recent latents, $\latenthist_t:=(\latent_{t-m}, \hdots, \latent_{t})$.
Given memory, history, and an $h$-step action plan $\acttraj_t$, the WM predicts $h$ \textit{future latents}:
\begin{equation}
    \latentpredtraj_t \sim \dyn(\cdot \mid \latentmem_t,\latenthist_t,\acttraj_t),
\end{equation}
where 
$\latentpredtraj_t:=\hat{\latent}_{t+1:t+h+1}$ is the $h$-step future.
We also train a reward model that scores the predicted latent's alignment with the language instruction:
$\mathbf{\hat{r}}_t \sim \reward(\cdot \mid \latentpredtraj_t,\lang)$ where $\mathbf{\hat{r}}_t := \hat{r}_{t+1:t+h+1}$. 
To enable iterative calls to the visuomotor policy, we use a pretrained decoder to obtain future observations (camera views and proprioceptive state), 
$\mathbf{\hat{\obs}}_t \sim \dec(\latentpredtraj_t)$, where $\mathbf{\hat{\obs}}_{t} := \hat{\obs}_{t+1:t+h+1}$ are the $h$-step future observations. The final prediction, $\hat{\obs}_{t+h+1}$, is fed back to the policy to generate the next action chunk.

\subsection{Key Design Decisions for High Fidelity, Temporally Consistent World Model Generation}
\label{sec:key_designs_world_model}

\para{Multi-View Camera Prediction}
Although using multiple views (e.g., wrist and external cameras) is common practice when designing visuomotor robot policies for manipulation to handle partial observability and capture finer-grained object--gripper interactions \cite{intelligence2025pi_}, many WMs only predict a single view~\cite{quevedo2025worldgymworldmodelenvironment, gao2026dreamdojo}. 
Following ~\citep{guo2026ctrlworld, jiang2025enerverse, wang2025vggt}, \ours predicts both external and wrist-camera observations. The increased information provided by multiple views helps with \textit{(ii)} consistency by helping handle occlusions during manipulation. Each view $I_t^j$ is encoded into $H \times W$ patch tokens using the pretrained Stable Diffusion 3 VAE encoder~\citep{esser2024scaling}. We project the proprioceptive state $q_t$ to the same token dimension and obtain $\latent_t$ by concatenating patch tokens and the proprioceptive token.

\para{Proprioceptive State Prediction}
In addition to future visual latents, \ours also predicts future proprioceptive states. We find that explicitly predicting the robot's configuration (rather than just visual observations like Ctrl-World~\citep{guo2026ctrlworld}) is critical to handle contact-rich manipulation of deformable objects, where knowing the precise position of the arm and width of the gripper is often required.

\para{Sparse Memory and Short-Term History}
Temporal \textit{(ii)} consistency across WM generations requires the WM to understand both what changes and what stays the same across an interaction. This is particularly challenging in manipulation, where occlusions and wrist camera viewpoint changes can cause objects and parts of the background scene to leave and enter the robot's FOV. In response, \ours builds upon \citep{guo2026ctrlworld} and conditions on two sets of observations when generating futures: a long-term, sparse \textit{memory}, and a short-term \textit{history}. In particular, memory $\latentmem_t := (\dots, \latent_{t-2k}, \latent_{t-k})$ includes every $k$th encoded observation to help capture longer-term context, while history $\latenthist_t:=(\latent_{t-1}, \latent_{t})$ includes the last two frames to capture the shorter-term consequences of actions.

\para{Latent Dynamics Model}
The latent dynamics model $\latentpredtraj_t \sim \dyn(\cdot \mid \latentmem_t,\latenthist_t,\acttraj_t)$ predicts future latent states conditioned on memory, history, and a candidate action plan. To balance \textit{(i)} fidelity with \textit{(iii)} efficiency, \ours adopts an efficient 2D transformer architecture following~\cite{hafner2025training, quevedo2025worldgymworldmodelenvironment}, with $L$ dynamics blocks composed of spatial attention and causal temporal attention. At each prediction step, the model conditions on latent tokens, action tokens, and flow timestep embeddings to autoregressively generate an $h$-step chunk. For stable training, each block uses RMSNorm~\citep{zhang2019root}, RoPE~\citep{su2024roformer}, QKNorm~\citep{henry2020query}, and SwiGLU feed-forward layers~\citep{shazeer2020glu} (see ~\ref{app:implementation_training} for more details).%

\para{Training Objective}
\label{sec:training_objective}
Similar to \citep{hafner2025training}, we train our latent dynamics model $\dyn$ with a flow-matching loss ~\cite{lipman2022flow} to predict future latents. Let $x^1_t := \latent_{t+1:t+h+1}$ denote the ground-truth next $h$ latents and let $x^0_t \sim \mathcal{N}(0, I)$ denote a Gaussian noise vector of the same dimension. Next, we define $x_t^{\tau} = \tau x_t^1 + (1-\tau)x_t^0$, with $\tau \in [0,1)$. Then, we train $\dyn$ to predict ``velocity'' $x^1_t - x^0_t$ by minimizing mean squared error: $\mathcal{L}^{\texttt{WM}}(\phi)=\mathbb{E}_{x^0_t, x^1_t, \tau}\left[\left\|
    (x_t^1 - x_t^0)
    -
    \dyn(\latenthist_t, \latentmem_t, \acttraj_t, x^{\tau}_t, \tau)
    \right\|_2^2\right]$. To improve long-horizon \textit{(ii)} consistency, we adopt Diffusion Forcing~\citep{chen2024diffusion}, which trains the latent dynamics model with independently sampled noise levels across future timesteps. We also use SPRINT blocks~\citep{park2025sprint}, which aggressively drop patch tokens in the latents to improve \textit{(iii)} efficiency.

\subsection{Accelerating World Model Inference Speed}
\label{sec:faster_inference}
For diffusion transformer-based WMs~\cite{guo2026ctrlworld,hafner2025training,quevedo2025worldgymworldmodelenvironment} like \ours, latency is a product of both \textit{(a)} the forward pass through the model and \textit{(b)} iterative denoising. Thus, \textit{(iii)} efficient generation requires tackling both of these concerns in tandem. We reduce cost \textit{(a)} via the use of KV caching to memory and history tokens across denoising steps. We reduce cost \textit{(b)} by adjusting the denoising process. In particular, building on diffusion forcing ~\cite{chen2024diffusion}, we use a progressive noise schedule. Rather than using a linear schedule like in~\cite{guo2026ctrlworld,hafner2025training}, \ours adopts a cosine schedule for higher \textit{(i)} fidelity generation.

To further increase \textit{(iii)} efficiency to the level required for test-time planning, we post-train \ours with a rectified flow objective~\cite{liu2022flow} to enable high-quality generation within a few forward passes. In particular, we first generate a high-quality latent trajectory using the denoising process, before using it as a target for secondary distillation step. See Appendix~\ref{app:inference_details} for more implementation details.

\subsection{Accurate and Efficient Value Estimation from the World Model}
\label{sec:value_estimate}

\para{Reward Model} To enable \textit{(iii)} efficient scoring of a proposed action chunk without needing to \textit{(a)} decode a latent into an image and \textit{(b)} feed it to an external VLM judge model~\cite{guo2026vlaw,quevedo2025worldgymworldmodelenvironment}, we distill the scores produced by an off-the-shelf reward model into a lightweight reward head that operates directly on latent states and language instruction $\lang$. The reward head $\reward$ aggregates latent tokens with AdaPool~\cite{brothers2026robust}, followed by MLP layers. We train $\reward$ with a simple mean squared error objective.

\para{Critic}
To support truncated-horizon rollouts with the WM, \ours learns a critic network $\critic$ that estimates the value beyond the imagined horizon. The critic shares the same latent-space design as the reward model and is trained with an MSE objective to predict bootstrapped $\lambda$-returns~\citep{sutton1998reinforcement}. Given latent rewards from $\reward$, the target is defined recursively as
$\mathbf{v}_t^{\lambda}
=
\reward(z_t,\lang)
+
\gamma \Big(
(1-\lambda)\critic(z_{t+1},\lang)
+
\lambda \mathbf{v}_{t+1}^{\lambda}
\Big)$, $\mathbf{v}_{t+k}^{\lambda} = \critic(z_{t+k},\lang)$.
The critic is then trained by minimizing $\mathcal{L}^{\texttt{critic}}(V)
=
\left\|
\critic(z_t,\lang) - \mathbf{v}_t^{\lambda}
\right\|_2^2$.

\subsection{Downstream WM Applications: Evaluation, Improvement, Planning}
\label{sec:downstream_application}

By satisfying the desiderata of \textit{(i)} fidelity, \textit{(ii)} consistency, and \textit{(iii)} efficiency simultaneously, \ours can support the downstream capabilities of evaluation, improvement, and planning.

\para{Policy Evaluation}
For policy evaluation, we take recorded action trajectories from real-world rollouts and execute them open-loop inside \ours, recording predicted reward values along the way. We focus on long-horizon tasks that sometimes require 40+ iterative evaluations of \ours's latent dynamics model, underscoring the importance of temporal \textit{(ii)} consistency and \textit{(iii)} efficiency.

\para{Policy Improvement} \label{method:policy_improvement}For policy improvement, we sample a $h$-step action chunk from the policy and forward simulate inside the WM $K$ times for a total of $H=Kh$ timesteps, leveraging \ours's \textit{(i)} fidelity and \textit{(ii)} consistency. After doing this $B$ times from the same initial observation $z_t$, we collect batch of rollouts $\{(\latent_t,\action_{t:t+H-1}^b,\hat{\latent}_{t+1:t+H}^b)\}_{b=1}^B$. We then compute a Monte-Carlo estimate of the $H$-step \textit{advantage} along each rollout: $\hat{A}_t^b
=
\sum_{\ell=1}^{H}
\gamma^{\ell-1}
\reward(\hat{\latent}_{t+\ell}^b,\lang)
+
\gamma^{H}
\critic(\hat{\latent}_{t+H}^b,\lang)
-
\critic(\latent_t,\lang).$ If the highest-scoring rollout in the batch (i.e., $\label{eq:best_imagined_sample}
b^\star = \arg\max_{b \in \{1,\ldots,B\}} \hat{A}_t^b$) has an advantage value above some small, positive threshold (i.e., $\hat{A}_t^{b^\star} > \epsilon_{\mathrm{adv}}$), we distill it into the base policy. This advantage-based filtering prevents the policy from being updated at states where all $H$-step sampled plans are predicted to be worse than the current expected behavior of the policy~\cite{jain2026a,anthony2017thinking}.

\para{Test-time Planning}
\label{sec:method_tts}
We adopt a single-chunk, best-of-$N$ \cite{lightman2023letsverifystepstep} approach to test-time scaling that doesn't involve iteratively calling the latent dynamics model. In particular, given the current observation and instruction, we sample $B$ candidate action chunks from the policy, imagine their outcomes with the world model, and execute the one with the highest advantage estimated with latent reward and critic heads. \ours's \textit{(iii)} efficiency (both in terms of the speed of the latent dynamics model and ability to evaluate a candidate action sequence without needing to call an external VLM judge via the the use of the reward head) are critical to unlocking this test-time scaling capability.

\section{Experimental Setup}
\label{sec:experiment-setup}

\para{Base Policy \& Hardware}
Our base policy is $\pi_{0.5}$~\citep{intelligence2025pi_}, a state-of-the-art vision-language-action (VLA) policy trained on the DROID dataset~\cite{droid}. 
We follow the DROID hardware setup and use a single Franka Emika Panda manipulator, two external Zed 2i cameras mounted on the left and right sides of the workspace, and a wrist-mounted Zed Mini camera (see Figure~\ref{fig:hardware-and-tasks} in Appendix). 
The $\pi_{0.5}$ VLA policy and our \ours world model use only the right camera view and the wrist camera\footnote{We setup all three cameras because our main world model baseline \cite{guo2026ctrlworld} uses all three views.}. 

\para{Datasets \& Tasks}
To align the world model with the data distribution of the base policy, we first pre-train the \ours world model on the DROID dataset and then fine-tune it on our real-world setup.
We collect data to fine-tune the world model $\realdataset^{\text{FT}}$ by running $\pi_{0.5}$ for five real-world manipulation tasks, with 50 rollouts per task. 
We also collect an additional 20 rollouts per task as evaluation data $\realdataset^{\text{val}}$. 
We select tasks such that the base policy achieves at least $20\%$ success rate while spanning a range of capabilities from rigid object pick-and-place to deformable object manipulation and dynamic manipulation. 
Specifically, our tasks are: \textbf{\textit{Stack Bowls}} (stack one bowl on another); \textbf{\textit{PnP Bag}} (place a deformable chip bag onto a plate); \textbf{\textit{PnP Marker}} (reorient a marker and insert it into a cup); \textbf{\textit{PnP Towel}} (place a soft towel into a basket); and \textbf{\textit{Pour Beans}} (pour a cup full of coffee beans into a bowl). Details on each task can be found in Appendix~\ref{appendix:task_details}.

\para{World Model Training} \ours is a 928M parameter model. We pretrain on the DROID dataset~\cite{droid} for 1M steps with a batch size of 32 and learning rate of $1e^{-4}$ on $4\times H100$ GPUs for 10 days. 
For training the reward model and critic on top of \ours's latents, we annotate the DROID dataset with progress-rewards obtained from Robometer~\cite{liang2026robometer} (reduced by 1 to get negative rewards). 
During world model finetuning, the model is updated with a lower learning rate of $2e^{-5}$ for 16k steps on our collected task data.
The resulting model is used for policy evaluation, policy finetuning, and test-time planning. 
Like prior work~\cite{guo2026ctrlworld}, we downsample the steps by 3 to use frequency of 5Hz for world model imagination.
We represent actions as the joint position difference between two timesteps to match the action space of the $\pi_{0.5}$ policy. 
We learn an additional joint-velocity-to-position action adapter to convert between the action spaces for data generation and test-time planning~(see~\ref{app:action_space}).

\section{Results}
\label{sec:experiment}

We first study the performance of the \ours world model in isolation (Sec.~\ref{subsec:results-wm-comparison}) and then in the downstream use-cases of policy evaluation, improvement, and test-time planning (Sec.~\ref{subsec:results-downstream}). 

\subsection{\ours Pareto-Dominates leading Manipulation World Models}
\label{subsec:results-wm-comparison}

We start by comparing the performance of \ours pre-trained only on the DROID dataset to leading multi-view manipulation world model. 
Ctrl-World~\cite{guo2026ctrlworld} is a 1.5B-parameter diffusion model trained on the DROID dataset and is initialized from a pretrained SVD checkpoint~\cite{blattmann2023stable}.

\para{Setup \& Metrics} 
We evaluate both models on a validation split of the DROID dataset (256 trajectories) and an out-of-distribution dataset $\realdataset^{\text{val}}$ collected using $\pi_{0.5}$ VLA (100 trajectories).
For each trajectory, the models are rolled out autoregressively to generate 10s long sequences where each generation predicts the outcome of 15-step action chunks~(1s) jointly. 
Following prior evaluations \cite{guo2026ctrlworld}, we measure the visual fidelity of the decoded generations using  FID~\cite{heusel2017gans}, and FVD~\cite{unterthiner2018towards} computed with the ground-truth videos. More metrics are detailed in the Appendix~\ref{app:world_model_eval}.

\begin{figure}[t]
\centering
\begin{minipage}{0.55\linewidth}
\centering
\setlength{\tabcolsep}{4.2pt}
\renewcommand{\arraystretch}{1.05}
\footnotesize
\begin{tabular}{l c cc cc c}
    \toprule
    & & \multicolumn{2}{c}{\textit{Exterior}} & \multicolumn{2}{c}{\textit{Wrist}} & \textbf{Time}\\
    \cmidrule(lr){3-4} \cmidrule(lr){5-6}
    \textbf{Method} & \textbf{NFE}
    & \textbf{FID} $\downarrow$ & \textbf{FVD} $\downarrow$
    & \textbf{FID} $\downarrow$ & \textbf{FVD} $\downarrow$
    & \textit{(s)} $\downarrow$ \\
    \midrule
    \rowcolor{pastellavender}
    \multicolumn{7}{c}{\textbf{\textit{\textcolor{slategray}{DROID (val)}}}} \\
    \ctrlworld & 16 & 26.09 & 78.73 & 33.83 & 195.37 & 14.65 \\
               & 50 & 22.44 & 55.05 & 25.32 & 91.77  & 42.33 \\
    \midrule
    \ours      & 16 & \cellcolor{lightblue}10.20 & 27.83 & 21.50 & 90.72 & \cellcolor{lightblue}\textbf{4.78} \\
               & 50 & \cellcolor{lightblue}\textbf{9.51} & \cellcolor{lightblue}\textbf{26.54} & \cellcolor{lightblue}\textbf{16.75} &
  \cellcolor{lightblue}\textbf{66.89} & 14.25 \\

    \midrule
    \rowcolor{pastelmint}
    \multicolumn{7}{c}{\textbf{\textit{\textcolor{slategray}{Task data (OOD)}}}} \\
    \ctrlworld & 16 & 36.16 & 139.54 & 38.76 & 277.13 & 14.65 \\
               & 50 & 31.44 & 91.48  & 33.47 & 145.86 & 42.33 \\
    \midrule
    \ours      & 16 & \cellcolor{lightblue}23.95 & \cellcolor{lightblue}88.27 & 30.77 & 184.62 & \cellcolor{lightblue}\textbf{4.78} \\
               & 50 & \cellcolor{lightblue}\textbf{23.48} & \cellcolor{lightblue}\textbf{87.03} & \cellcolor{lightblue}\textbf{27.37}
  & \cellcolor{lightblue}\textbf{145.04} & 14.25 \\
    \bottomrule
  \end{tabular}

\captionof{table}{We report FID and FVD on DROID(val) and OOD Task datasets and inference time at different NFEs. \ours pareto dominates Ctrl-World on fidelity vs inference budget (NFE and inference time).}
  \label{table:video_metrics}
\end{minipage}
\hfill
\begin{minipage}{0.42\linewidth}
\centering
\includegraphics[width=\linewidth]{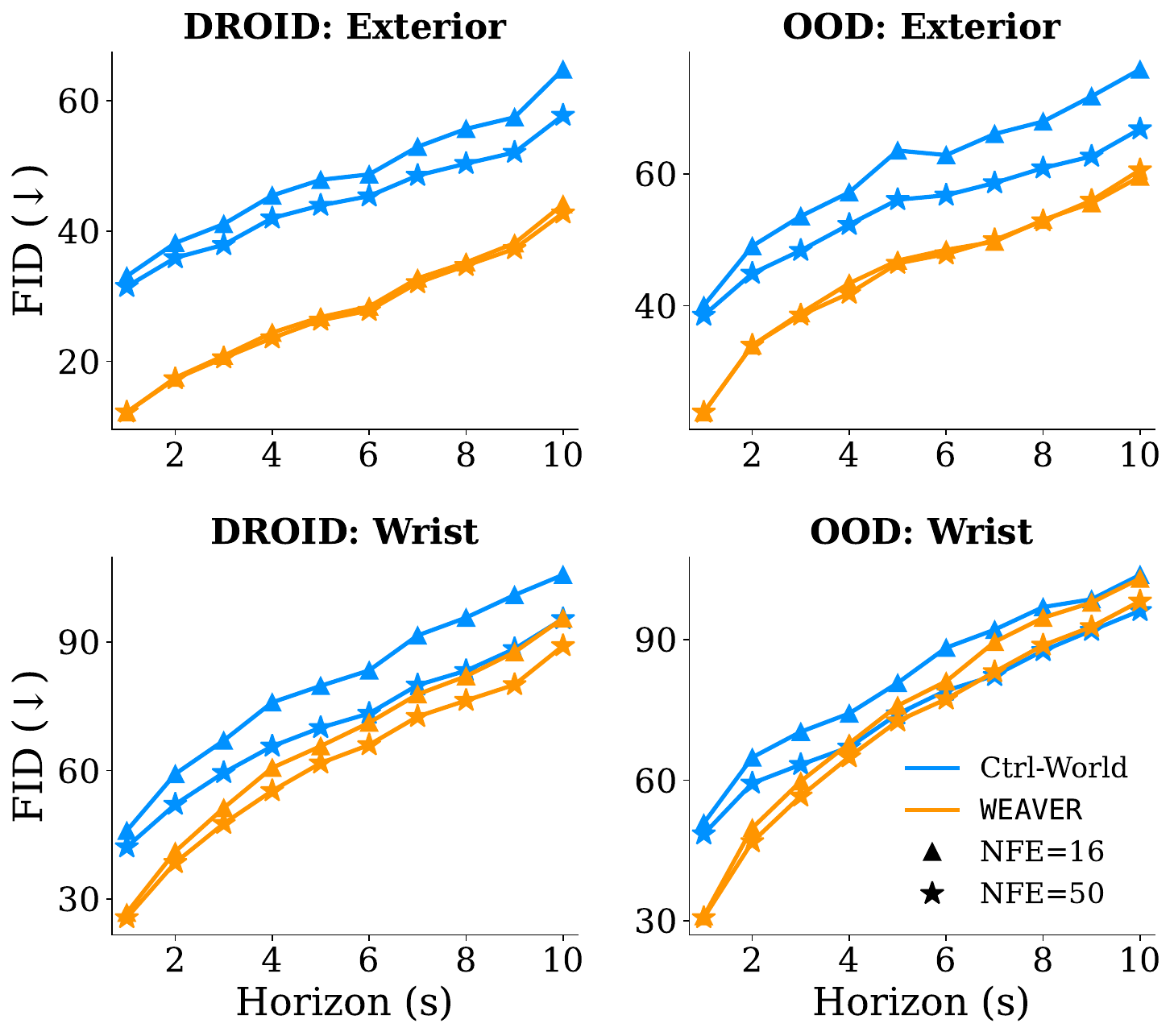}
\captionof{figure}{We report FID at various horizon lengths and find that \ours is consistently better at long-horizon rollouts.}
\label{fig:fid_over_time}
\end{minipage}

\end{figure}

\para{Results: Perceptually High Fidelity Generations} Table~\ref{table:video_metrics} compares FID and FVD results for \ours and Ctrl-World on different evaluation datasets. 
\ours outperforms the Ctrl-World while having lower inference time.
As we decrease the number of function evaluations (NFE) to decrease the latency, we find that 
the quality of Ctrl-World decreases more significantly than \ours; both models incur the highest error when predicting wrist camera viewpoints. 
We provide additional results to compare NFEs~(Appendix~\ref{app:quantitative_results}), different noise schedules~(Appendix~\ref{app:noise_schedule_result}) and inference speedup obtained with KVcaching~(Appendix~\ref{app:kv_cache_results}). 

\para{Results: Higher Quality at Long Horizon}
We next measure how the perceptual quality of the world model's imaginations are influenced by long-horizon predictions. 
For both world models, we generate rollouts with long (150-step or 10s) action sequences and measure the FID for each predicted video of the 15-step interval to estimate the generation quality with time horizon.
As shown in Fig.~\ref{fig:fid_over_time}, on the DROID dataset, we find that \ours maintains consistently lower FID compared to \ctrlworld even as inference budgets are reduced from 50 to 16 NFE. 
On the OOD dataset, \ours maintains the performance gap on exterior-view and has comparable performance on the wrist-view.

\begin{figure}
    \centering
    \includegraphics[width=\linewidth]{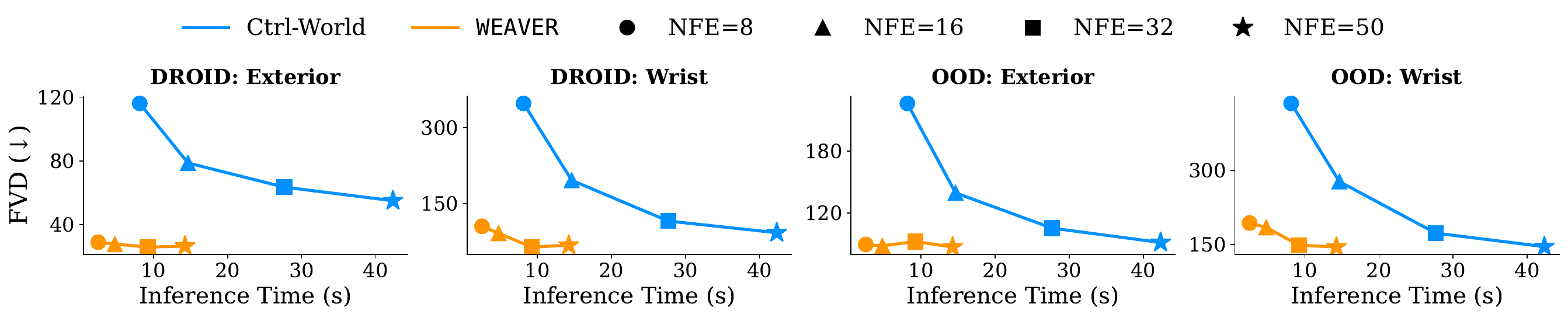}
    \caption{We present FVD vs inference time (in seconds) for \ours and \ctrlworld across views on DROID (val) and OOD task datasets. We find that \ours pareto-dominates the leading method Ctrl-World at different NFE while using upto $16\times$ less inference time.}
    \label{fig:inference_time}
    \vspace{-1em}
\end{figure}

\para{Results: \ours Pareto-Dominates Inference Speed vs. Quality}
Next, we study how the generation quality is influenced by a fixed inference time budget as measured by NFEs and the inference time to generate 10s chunk on a single H100 GPU. 
In Fig.~\ref{fig:inference_time}, 
we see that \ours significantly outperforms \ctrlworld at NFEs from 8,16,32,50 while enjoying significantly lower inference speeds (e.g., 30-50s with Ctrl-World vs. 10-30s for \ours). 
By pareto-dominating Ctrl-World, \ours unlocks faster evaluation and planning as we explore below in Section~\ref{subsec:results-downstream}. 

\begin{figure}
    \centering
    \includegraphics[width=\linewidth]{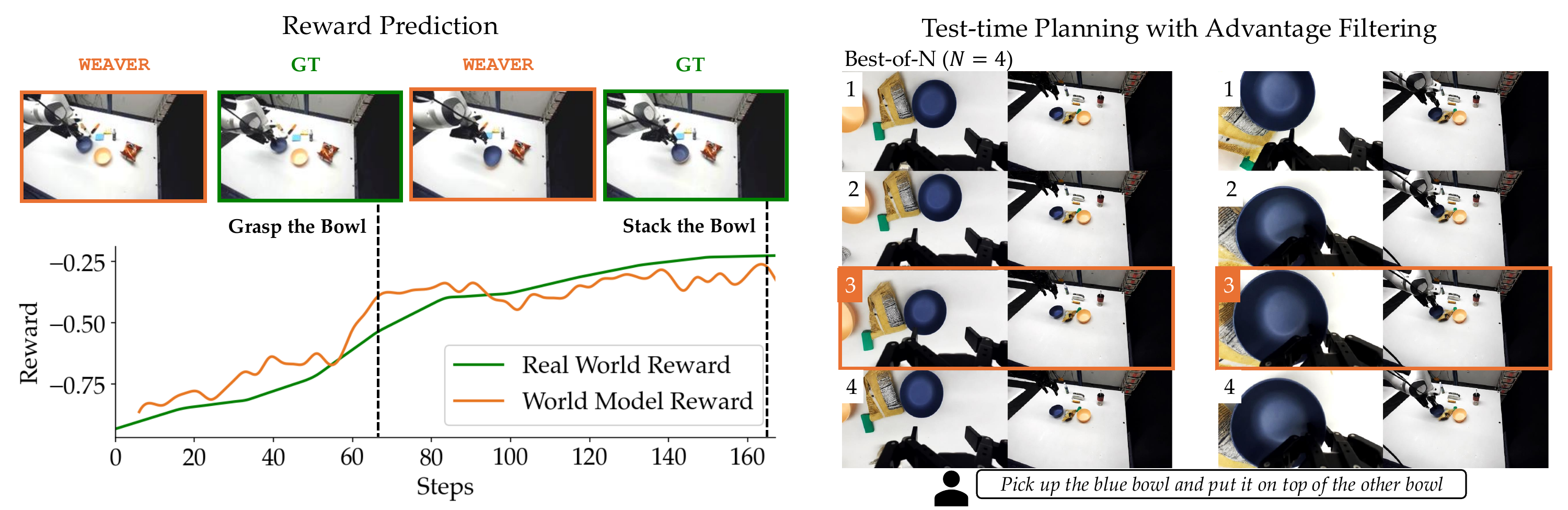}
    \caption{\textbf{Reward Prediction \& Test-time Planning with Advantage Filtering}.  (Left) Predicted rewards from \ours match the Robometer reward over trajectory.  (Right) The highlighted action sample is the one with the best advantage value and the best outcome in \ours's imagination.}
    \label{fig:reward_plot}
    \vspace{-1em}
\end{figure}

\para{Results: Latent Reward Prediction Accuracy}
Finally, we compare \ours's latent reward prediction to the reward labels from RoboMeter~\cite{liang2026robometer}, evaluated on real held-out trajectories. Fig.~\ref{fig:reward_plot} shows the predicted reward for a rollout of the \textbf{PnP Stack} task; \ours correctly imagines key events such as grasping and stacking and the reward of the imaginations correlates with the ground-truth RoboMeter reward. 
In the right panel of Fig.~\ref{fig:reward_plot}, we see that the advantage computed with the predicted reward is also able to distinguish different outcomes of the action samples. 
This is a promising indicator that \ours and it's latent reward are suitable for filtering synthetic data in Sec.~\ref{subsec:result-policy-eval} and test-time planning in Sec.~\ref{subsec:result-test-time-plan}.

\subsection{\ours Enables Effective Evaluation, Improvement and Planning} 
\label{subsec:results-downstream}

Thus far, we have validated that \ours effectively balances \textit{(i)} fidelity, \textit{(ii)} long-horizon consistency, and \textit{(iii)} efficient generation. 
Next, we turn to the downstream uses of a world model: policy evaluation (Sec.~\ref{subsec:result-policy-eval}), policy improvement (Sec.~\ref{subsec:result-policy-improve}), and test-time planning (Sec.~\ref{subsec:result-test-time-plan}).

\subsubsection{\ours Enables Effective Policy Evaluation that Tightly Correlates with Reality}
\label{subsec:result-policy-eval}

First, we evaluate whether \ours can serve as a learned simulator for offline policy evaluation, reducing the need for costly real-world rollouts. 

\para{Setup}
Given an initial real observation, $\obs_0$, and action sequence, $\acttraj_t \sim \policy_\theta(\cdot \mid \obs_0, \lang)$, we autoregressively generate imagined observations and estimate policy performance from the resulting rollout. 
We compare three world models: \ctrlworld pretrained on DROID~\citep{guo2026ctrlworld}, \ours pretrained on DROID, and \oursft finetuned on $\mathcal{D}_{\mathrm{real}}^{\mathrm{FT}}$. To test robustness of the world models across base policy quality, we evaluate each model on rollouts from both the base $\pi_{0.5}$ policy and a finetuned policy. %

\para{Metrics}
Following prior work~\citep{yin2026playworld}, we measure how well performance of generated rollouts correlates with real-world performance by comparing human-labeled binary success rates on imagined rollouts with real success rates on $\mathcal{D}_{\mathrm{real}}^{\mathrm{val}}$, averaged over 20 trials per task. 
We report Pearson Correlation coefficient~\citep{pearson1920notes} and maximum matrix ranking violation (MMRV)~\citep{team2025evaluating} (see Appendix~\ref{app:policy_evaluation}).

\begin{figure}
    \centering
    \includegraphics[width=\linewidth]{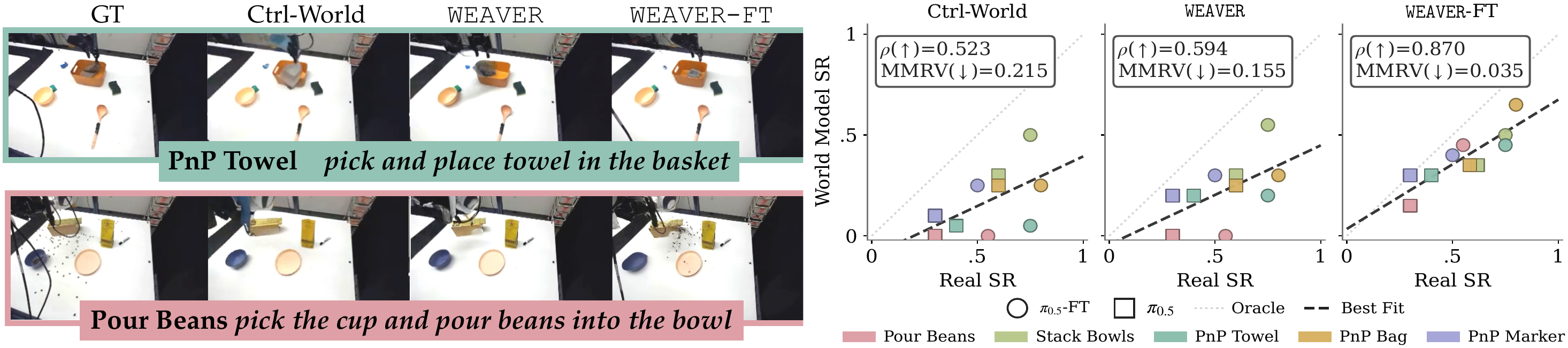}
    \caption{\textbf{Policy Evaluation}. We compare performance across different policies and world models. (Left) For PnP Towel, only \ours and \oursft accurately imagine the towel inside the basket. For Pour Beans, only \oursft captures the beans scattering across the table. (Right) Evaluation inside \oursft attains an impressively high correlation of success rate with the real world.}
    \label{fig:policy_eval}
    \vspace{-0.2cm}
\end{figure}

\para{Results}
Fig.~\ref{fig:policy_eval} shows that pretrained world models tend to \textit{underestimate} policy performance, but \ours achieves better agreement with real rollouts than \ctrlworld, with higher Pearson correlation and lower MMRV. This setting is challenging because rollouts can last up to 40 seconds and require accurate long-horizon prediction. 
The pouring task is particularly difficult for pretrained models, likely because granular dynamics are underrepresented in DROID and inherently hard to model. 
After finetuning, \oursft substantially improves evaluation accuracy, increasing Pearson correlations to $\rho=0.87$ and better matching real outcomes across policies of varying performance. 
The qualitative example on the left of Fig.~\ref{fig:policy_eval} further shows that \oursft predicts the \textbf{PnP Towel} and \textbf{Pour Beans} task outcomes more accurately than the baselines.

\subsubsection{\ours Enables Effective Policy Improvement Without Real Interactions}
\label{subsec:result-policy-improve}

Another desirable use of high-fidelity world models is synthetic data generation for policy improvement. 
We use the world model to sample and verify candidate action segments, then distill high-value imagined segments back into the policy~\cite{jain2026a,anthony2017thinking}.

\para{Setup}
To evaluate the utility of \ours towards improving policies, we explore various strategies to generate data for finetuning the policy:
(1) \textbf{Base Policy}: the original $\pi_{0.5}$ VLA trained on DROID; 
(2) \textbf{FT w/ Real Data}: we prune segments in real trajectories using advantage estimates, yielding 1,000 segments of 36-step action chunks per task; 
(3) \textbf{FT w/ Synthetic Data}: we sample multiple segments using the base policy and \ours, filter them based on predicted advantage values (Sec.~\ref{method:policy_improvement}), and retain 1,000 segments per task,
and (4) \textbf{FT w/ Mixed Data}: combine filtered real and synthetic datasets (2000 segments per task) (more details and results are presented in App.~\ref{app:policy_finetuning} \& ~\ref{app:policy_improvement}).

\begin{figure}[t]
    \centering
    \begin{subfigure}[t]{0.74\linewidth}
        \centering
        \includegraphics[width=\linewidth]{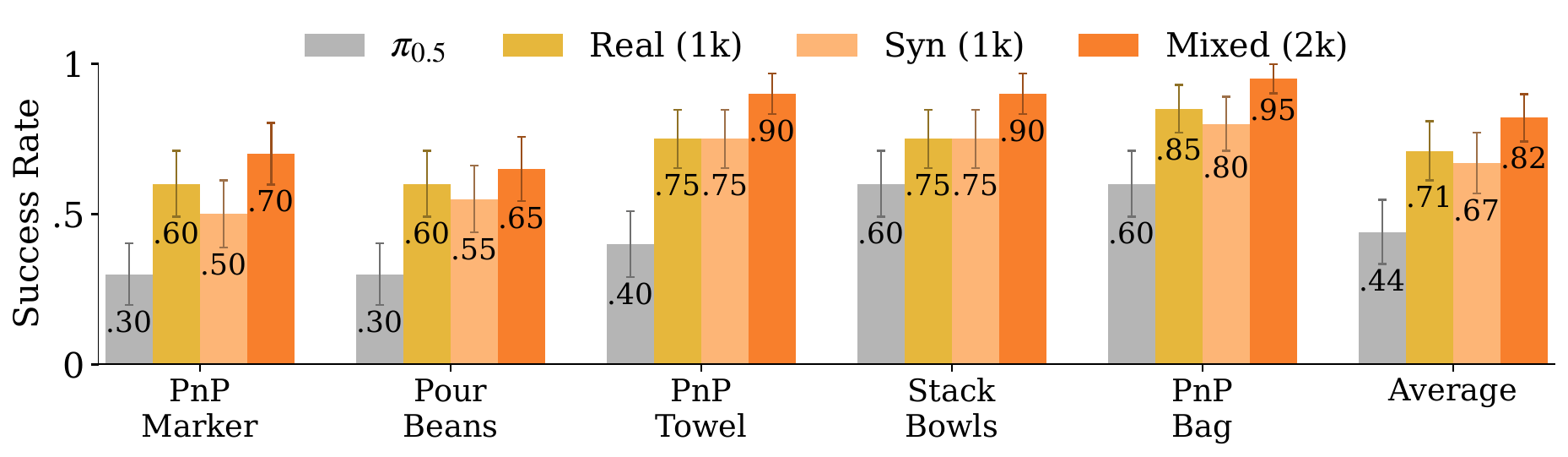}
      
    \end{subfigure}
    \hfill
    \begin{subfigure}[t]{0.24\linewidth}
        \centering
        \includegraphics[width=\linewidth]{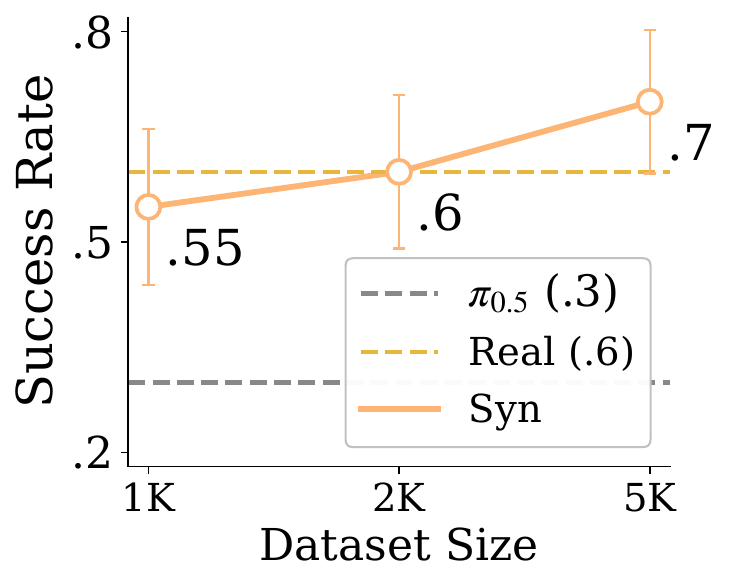}
      
    \end{subfigure}
    \caption{(Left) \textbf{Policy Improvement with Finetuning.} We finetune $\pi_{0.5}$ with multiple data sources and see that combining real and synthetic~(Syn) obtained with \ours outperforms other variants. (Right) \textbf{Data Scaling for Policy Improvement.} We ablate the number of segments in synthetic data for finetuning and report the success rate across 20 trials for the \textbf{Pour Beans} Task. }
    \label{fig:policy_improvement}
\end{figure}

\para{Results}
Fig.~\ref{fig:policy_improvement} shows that all finetuned policies substantially improve their success rate over the base policy. 
Notably, finetuning on synthetic data closely matches that on real data, with only a $4\%$ average performance gap. 
This indicates that out synthetic data is of such a high quality that it unlocks similar policy improvement to costly real world data.
Combining real and synthetic data further improves performance, increasing the average success rate by $11\%$ over real-data finetuning alone. 
These results suggest that imagined rollouts from the world model provide a useful source for distillation, reducing the need for costly real-world collection and manual filtering. 
Fig.~\ref{fig:policy_improvement} also shows improvements on contact-rich and dynamic manipulation tasks, such as more precise marker placement and bean pouring.
We further study synthetic data scaling on the \textbf{Pour Beans} task by varying the number of imagined segments from 1,000 to 2,000 and 5,000. Fig.~\ref{fig:policy_improvement}~(right) shows that policy performance improves consistently with more synthetic data, eventually \textit{exceeding the performance obtained from real-data finetuning alone}.

\begin{figure}
    \centering
    \includegraphics[width=\linewidth]{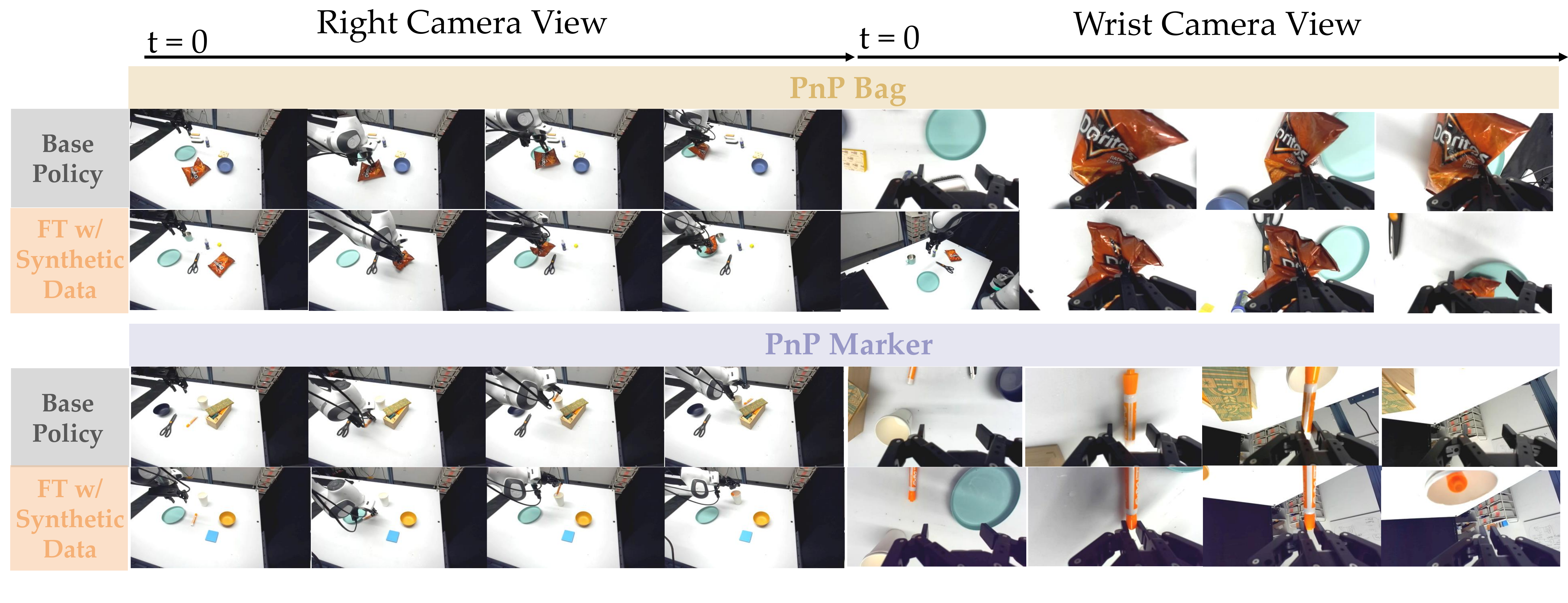}
    \caption{\textbf{Policy Improvement Results}. We present real rollouts from the base policy and the policy finetuned with synthetic data. Finetuning on synthetic data generated by \ours leads to improved policy performance and more successful task execution compared to the base policy.}
    \vspace{-0.4cm}\label{fig:policy_improvement_main}
\end{figure}
\subsubsection{\ours Enables Test-Time Planning by Balancing Inference Speed and Quality}
\label{subsec:result-test-time-plan}

Finally, test-time search requires evaluating multiple action sequences before execution, making inference speed a key bottleneck. 
In contrast to planning in the the image space using reconstruction and VLM-as-a-judge~\cite{guo2026vlaw,sharma2026world}, \ours plans in the latent space for greater efficiency~\cite{hafner2021mastering,jain2026a}.

\begin{wrapfigure}{r}{0.48\textwidth}
    \vspace{-20pt}
    \centering
    \includegraphics[width=0.46\textwidth]{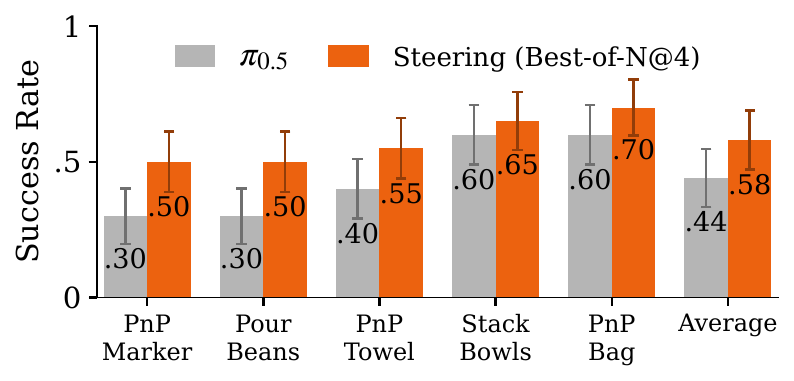}
    \vspace{-6pt}
    \caption{We demonstrate test-time steering with \ours outperforms the base policy $\pi_{0.5}$ by 14\% when averaged across all five tasks.}
    \vspace{-25pt}
    \label{fig:test-time-sr}
\end{wrapfigure}

\para{Setup} We use $\pi_{0.5}$ as the base policy and sample a batch of action chunks. For each chunk, \ours imagines latents of future states and evaluates the advantage using the reward and critic heads.
This reduces the cost of decoding predicted observations and querying external VLM judges. 
Following the policy-improvement setup from Sec.~\ref{subsec:result-policy-improve}, we evaluate test-time planning on five tasks and compare against the base policy. 
We use $B=4$ parallel samples and an imagination horizon of $h=12$, balancing planning quality and latency.

\para{Results}
We report task success rate and the inference-time breakdown in the test-time planning pipeline.
Fig.~\ref{fig:test-time-sr} shows that advantage-based selection steers the policy toward successful behaviors.
Test-time planning improves the average success rate by $15\%$ over the base policy, with maximum gain up to $20\%$. The improvement is larger when the base policy is weaker, although it remains smaller than direct finetuning because planning is limited to a single action chunk and must operate under latency constraints. Table~\ref{tab:inference_time} in Appendix~\ref{app:inference-time-latency} shows dynamics prediction remains the main computational bottleneck. Nevertheless, \ours is about $20\times$ faster than \ctrlworld inference pipeline~\citep{guo2026ctrlworld} on an RTX A6000 Ada GPU, and batched sampling scales sublinearly with the number of candidates, showing that our inference optimizations make world-model-based test-time planning practical for real-time manipulation.

\section{Conclusion}

We introduce \ours: a World Model for manipulation that achieves \textit{(i)} high fidelity, \textit{(ii)} is temporally coherent, and \textit{(iii)} generates efficiently. Across tasks, \ours shows strong correlation  ($\rho=0.870$) with real-world success rate for evaluation, improves the success rate of the $\pi_{0.5}$ policy by $38\%$ \textit{without} any real-world interaction, and unlocks test-time planning $5-10\times$ faster than Ctrl-World~\cite{guo2026ctrlworld}.

\para{Limitations}
While \ours unlocks the potential of large-scale world models for manipulation, several limitations remain. First, visual world models observe only a partial view of the underlying state, and tactile sensing may be necessary to resolve ambiguities. Second, incorporating physics priors could improve performance on tasks involving deformable-object manipulation. Third, generation latency currently limits test-time planning to short-horizon reasoning over a single action chunk. Finally, reward supervision from RoboMeter can be noisy, motivating the development of better reward models for failure prediction. We provide further discussion in Appendix~\ref{app:limitations}.

\para{Broader Impact} This work explores large-scale world models to improve the efficiency, safety, and scalability of robotic manipulation by reducing reliance on costly real-world interaction. Imagined rollouts can support policy evaluation, improvement, and test-time planning before execution, but inaccurate or biased predictions may lead to risky decisions that are particularly important in safety-critical domains like assistive robots. Responsible deployment therefore requires careful validation, uncertainty estimation, and safeguards against exploiting errors in learned world or reward models.

\section*{Acknowledgments} 
We would like to thank Jesse Zhang for helpful discussions about reward models and ROBOMETER.
AJ is supported by Fonds de Recherche du Quebec (FRQ) (DOI assigned: https://doi.org/10.69777/350253),
Calcul Quebec, and
Canada Excellence Research Chairs (CERC) program.
GKS is supported by a STTR grant. 
The research was enabled in part by computational resources provided by the Digital Research Alliance of Canada
(https://alliancecan.ca) and Mila (https://mila.quebec).
YW and AB were partially supported by the National Science Foundation (NSF) award $[\#2246447]$ and NSF CAREER award $[\#2441014]$]. The views expressed are those of the authors and do not necessarily reflect those of NSF.
\clearpage

\bibliographystyle{plainnat}
\bibliography{references}

\clearpage

\appendix

\setcounter{section}{0}
\renewcommand{\thesection}{A\arabic{section}}

\section*{Contents}
\startcontents[appendix]
\printcontents[appendix]{}{1}{}
\clearpage

\clearpage

\section{Robot Setup \& Tasks}

\subsection{Tasks Details}
\label{appendix:task_details}
\begin{figure}[ht!]
    \centering
    \includegraphics[width=\linewidth]{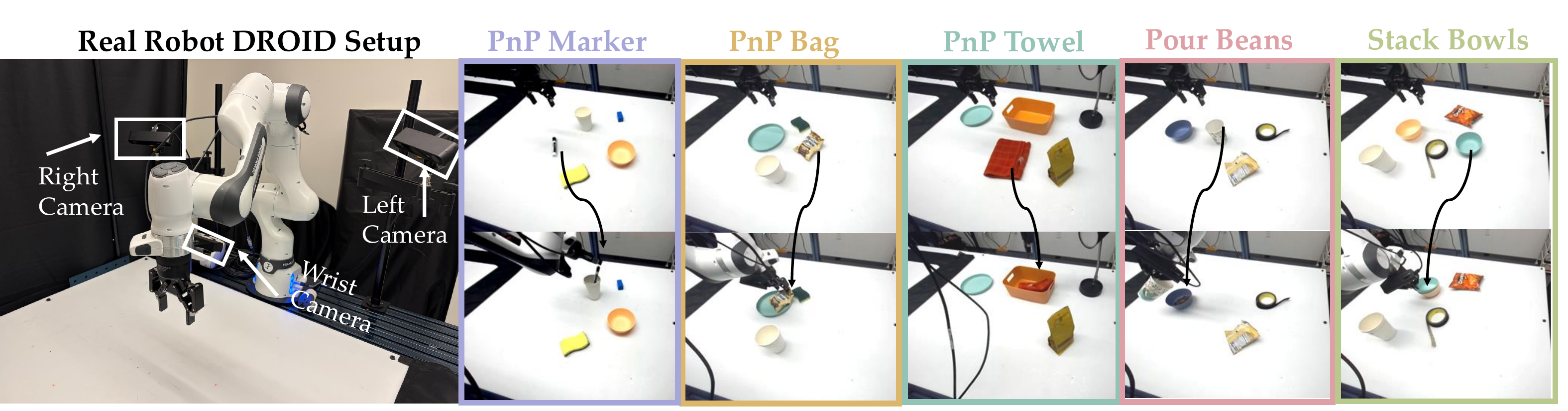}
    \caption{\textbf{Hardware setup and tasks.}On the left, it is the robot setup with cameras. On the right, it shows the five tasks with top row as initial state and bottom row as one of the goal configuration.}
    \label{fig:hardware-and-tasks}
\end{figure}

We collect real-world finetuning data from $\pi_{0.5}$ on five manipulation tasks, with 50 rollouts per task on our DROID setup as shown in Fig.~\ref{fig:hardware-and-tasks}. We select tasks for which the base policy achieves at least $20\%$ success, ensuring that the collected rollouts contain both successful and failed executions while remaining within the policy's competence. The tasks are designed to cover a diverse set of manipulation regimes, including rigid-object pick-and-place, deformable-object manipulation, and dynamic manipulation as shown in Fig.~\ref{fig:hardware-and-tasks}.

\textbf{\textit{Stack Bowls}} requires the robot to stack one bowl on top of another. Two bowls are randomly placed on the table, and the robot must place bowl $A$ on bowl $B$, where $A \in \{\text{blue}, \text{green}\}$ and $B \in \{\text{blue}, \text{green}, \text{pink}\}\setminus A$.

\textbf{\textit{PnP Bag}} requires the robot to pick up a bag of chips and place it on a green plate. We use two types of chip bags and randomly sample one in each episode. The bag is deformable, making the grasp outcome and object motion difficult to predict.

\textbf{\textit{PnP Marker}} requires the robot to pick up an Expo marker lying horizontally on the table and place it inside a container. The marker color is randomly selected from black and orange, and the target container is randomly selected from a paper cup and a blue mug. This task requires precise grasping and large end-effector reorientation to insert the marker vertically.

\textbf{\textit{PnP Towel}} requires the robot to pick up a towel and place it into a basket. We use two towel variants, a folded thick red kitchen towel and a thin gray square towel, and two basket variants, orange and blue. The task is challenging because the towel is deformable, and its resulting shape depends strongly on the grasp location and, for the folded towel, the number of layers grasped.

\textbf{\textit{Pour Beans}} requires the robot to pick up a cup containing coffee beans and pour them into a blue bowl. This task tests dynamic manipulation, as the granular motion of the beans is difficult to predict and successful execution requires accurate control of cup pose, pouring angle, and motion to avoid spilling outside the bowl.

\subsection{Action Space}
\label{app:action_space}
The $\pi_{0.5}$ base policy on the DROID setup outputs joint-velocity commands for control. To match this action representation, we define the action space of our world model in joint space, avoiding potential compounding errors from converting actions into alternative representations such as Cartesian space. However, we find that directly conditioning the world model on joint velocities leads to lower generation quality. Therefore, following prior work~\cite{guo2026ctrlworld}, we use a lightweight action adapter to convert joint velocities into joint positions.

For policy evaluation, because the joint positions at the end of each trajectory are already available, we directly use joint positions as inputs to the world model. During test-time planning, however, predictions must be made from the joint-velocity actions proposed by the policy. We therefore use the trained action adapter to predict the corresponding joint positions, and condition the world model generation on these adapted actions. The following section describes the details of the action adapter.
\subsubsection{Action Adapter}  
\paragraph{Overview.}                                                                          
  The action adapter is a lightweight feedforward module that bridges the world                                                  
  model's action representation (joint velocity commands and binary gripper                                                      
  signals) and the robot's observable state (absolute joint positions and gripper                                       width). Given the robot's current state and a chunk of $T{=}15$ actions                                                     
  produced by the world model, it predicts the resulting sequence of
  joint-position and gripper-position deltas, which are then integrated to obtain                                                
  future absolute states.                                                                                                        
   
  \paragraph{Input representation.}                                                                                              
  The model receives two groups of inputs:
  \begin{itemize}
    \item \textit{State token}: the current 7-DOF joint position concatenated
          with the current scalar gripper position, forming a $(7{+}1)$-dimensional                                              
          vector.
    \item \textit{Action tokens}: a chunk of $T$ joint-velocity commands                                                         
          ($T \times 7$) concatenated with $T$ gripper actions ($T \times 1$),                                                   
          flattened to $T(7{+}1)$ dimensions.                                                                                    
  \end{itemize}                                                                                                                  
  Both groups are concatenated into a single input vector of dimension                                                           
  $(7{+}1)(T{+}1) = 128$ (for $T{=}15$).
                                                                                                                                 
  \paragraph{Architecture.}
  The adapter is a three-layer MLP with hidden size 512 and SiLU activations:                                                    
  \[              
    f_\theta : \mathbb{R}^{128} \;\longrightarrow\; \mathbb{R}^{T \times 8}.                                                     
  \]                                                                                                                             
  The output is reshaped to $(T, 8)$ and split into predicted joint deltas                                                       
  ($T \times 7$) and predicted gripper deltas ($T \times 1$).                                                                    
                                                                                                                                 
  \paragraph{Normalization.}
  All continuous inputs and targets are min-max normalized to $[-1, 1]$ using                                                    
  per-dimension 1st/99th-percentile bounds computed from the training set, which
  is more robust to outliers than global min/max. Gripper action commands are                                                    
  binarized (${\geq}0.5 \mapsto 1$, otherwise $0$) prior to input, reflecting
  their discrete open/close semantics.                                                                                           
                                                                                                                                 
  \paragraph{Loss function.}
  The model is trained with a weighted MSE loss on the normalized delta targets:                                                 
  \[              
    \mathcal{L} \;=\; \mathcal{L}_{\text{joint}}
                  \;+\; \lambda_g\,\mathcal{L}_{\text{gripper}},                                                                 
    \qquad \lambda_g = 5.0.                                                                                                      
  \]                                                                                                                             
  The gripper is up-weighted because it has a much smaller dynamic range than the                                                
  joint dimensions and would otherwise be under-penalized relative to its importance in grasp and place predictions.                                                                                                           
   
  \paragraph{Inference.}                                                                                                         
  At test time the model predicts
  $(\hat{\Delta}_{\text{joint}},\,\hat{\Delta}_{\text{gripper}})$,
  denormalizes them, and integrates from the current state:                                                                      
  \[
    q_{t+k} = q_t + \hat{\Delta}_{\text{joint},k},                                                                               
    \qquad                                                                                                                       
    g_{t+k} = g_t + \hat{\Delta}_{\text{gripper},k},
    \qquad k = 1,\ldots,T.                                                                                                       
  \]                                                                                                                             
   
  \paragraph{Training details.}                                                                                                  
  The adapter is trained for 15 epochs using Adam
  ($\text{lr}{=}10^{-4}$, batch size 128) on 50 hours of proprioceptive
  teleoperation data. Each training sample consists of a randomly drawn window of                                                
  $T{+}1$ consecutive timesteps from an episode; the first timestep provides the                                                 
  current state and the remaining $T$ timesteps provide the action chunk and                                                     
  delta targets.  
\section{Implementation Details}
\label{app:implementation_training}
\subsection{Architecture Details}
We use the VAE from Stable Diffusion 3~\cite{esser2024scaling} to encode 190$\times$32 image frames from camera views into the latent space.
Our efficient transformer architecture is a 32-layer transformer with 1536 hidden dimensions and 16 attention heads.
Each layer comprises of a spatial layer that attends to all the patches in $z_t$ and a causal temporal layer to attend over patches from prior observations.
The actions and proproceptive states are normalized using statistics obtained with the training dataset of DROID.
We obtain the reward annotations from Robometer~\cite{liang2026robometer} and use the progress rewards to train the reward head and the critic.
The reward and critic networks uses an AdaPool~\cite{brothers2026robust} layer to compress the tokens to a vector, and has MLP layers with this vector concatenated with the CLIP embedding~\cite{radford2021learning}~(provides a representation for the language instruction).

\subsection{Training Details} 
\label{app:training_details}
\para{\ours Pretraining} \ours has 928M paramters in total and is trained for 1M gradient steps on 4$\times$H100 GPUs for 10 days. The pretraining is done on the DROID dataset~\cite{droid}. We also maintain an exponential moving average~(EMA) of model weights during training with $\beta=0.9999$. We use a learning rate warmup for the initial 10000 steps and keep a constant learning rate of $1e^{-4}$ post warmup. We provide the hyperparameters in Table ~\ref{app:tab_hyperparams}.

\para{\ours Finetuning} For complex tasks like pouring that is underrepresented in the pretraining dataset, the world model is inaccurate at predictions. 
To mitigate covariate shift and improve generations, a potential solution is to finetune the world model with task dataset.
In this work, we finetune \ours on a small datasets of 250 trajectories (50 for each task) collected using the $\pi_{0.5}$ VLA. 
We finetune \ours for 16K gradient steps using a smaller learning rate of $2e^{-5}$. Other hyperparameters are similar to pretraining (as described in Table~\ref{app:tab_hyperparams}) and the training takes 6 hours on 4$\times$H100 GPUs.

\para{$\pi_{0.5}$ Finetuning} 
\label{app:policy_finetuning}We follow the original setup of the droid finetuning configuration in openpi to finetune our policy. We use the open-sourced pre-trained $\pi_{0.5}-$droid checkpoint as the base policy.             
  Normalization statistics are inherited from the original DROID checkpoint and                                                  
  held fixed throughout fine-tuning to preserve compatibility with the pretrained                                                
  observation encoder.                                                           
  All runs with dataset size smaller than 5000 segments use a batch size of 32, a peak learning rate of $2.5\times10^{-6}$                                                    
  with cosine decay and no warmup, and language instructions sourced from task
  annotations.                                                                                                              
  For datasets of size ${\geq}5{,}000$ trajectories, we fine-tune for
  10{,}000 gradient steps with a cosine decay over 10{,}000 steps and warmup steps of 1000 and peak learning rate of 2.5e-5.                                       
  For smaller datasets (1{,}000--2{,}000 trajectories), we reduce training to mitigate                                          
  overfitting. 

  \para{Reward Labeling}We label each trajectory with a per-frame progress reward using the Robometer evaluation server~\cite{liang2026robometer}.              
  For each episode, we extract frames from the recorded video and downsampled to 1\,fps using, selecting the right-camera view  from the DROID setup.                                                                      
  The sampled frames and the episode's language instruction are sent in a single forward pass to the Robometer eval server, which
   returns a per-frame \emph{progress prediction} $\hat{r}_t \in [0,1]$ representing the estimated fraction of task completion at
   frame $t$, along with an optional per-frame success probability $\hat{s}_t \in [0,1]$.
                                                                                                                                 
  Because frames are subsampled at 1\,fps, the resulting reward sequence is shorter than the original video.                     
  We realign rewards to the full video length by linear interpolation: letting $T_{\text{orig}}$ denote the original frame count
  and $T_{\text{sampled}}$ the number of inferred frames, we place sampled values at positions                                   
  $\{(T_{\text{orig}}-1)\,i/(T_{\text{sampled}}-1)\}_{i=0}^{T_{\text{sampled}}-1}$ and interpolate onto the integer grid $\{0, 1,
   \ldots, T_{\text{orig}}-1\}$.                                                                                                                  
  We choose \texttt{reward\_progress} (the interpolated progress signal) as our final reward annotation because it is more aigned with actual task outcome. We substract the reward progress by -1 to make the reward fall in $[-1,0]$ as labels for training.                                                                              
\begin{table}[h]
\centering
\begin{tabular}{l|c}
    \toprule
    \textbf{Name} & \textbf{Value} \\
    \midrule
    \multicolumn{2}{c}{\textbf{World Models}} \\
    \midrule
    Layers & 32 \\
    Heads  & 16 \\
    Embedding dimension & 1536 \\
    Head dimension & 96 \\
    SPRINT probability & 0.5 \\
    \midrule
    \multicolumn{2}{c}{\textbf{Reward Model and Critic}}  \\
    \midrule
    MLP layers & 2 \\
    Discount factor~$\gamma$ & .995 \\
    Return lambda~$\lambda$ & .95 \\
    \midrule
    \multicolumn{2}{c}{\textbf{Pretraining}} \\
    \midrule
    Batch size & 32 \\
    Batch length & 8 \\
    Memory frames~(p) & 6 \\
    Memory frame stride~(m) & 5 \\
    Optimizer & AdamW \\
    Proprioceptive State loss scale ($\mathcal{L}_{q}$) &  0.1 \\
    LR & $1e^{-4}$ \\
    Warmup steps & 10000 \\
    EMA decay & .9999 \\
    Training Steps & 1000000 \\
    \midrule
    \multicolumn{2}{c}{\textbf{Finetuning}} \\
    \midrule
    Batch size & 32 \\
    LR & $2e^{-5}$ \\
    Warmup steps & 200 \\
    EMA decay & .9999 \\
    Training Steps & 16000 \\
    \bottomrule
\end{tabular}
\vspace{4pt}
\caption{\textbf{Hyperparameters.} We present the list of hyperparameters used for training \ours.} \label{app:tab_hyperparams}
\end{table}
\subsection{Inference}
\label{app:inference_details}

\para{Inference noise schedules}
We evaluate several deterministic schedules to map discrete inference index $i \in \{0,\ldots,K\}$ to the noise level $k \in [0,1]$, where $K$ is the number of denoising steps. 
We describe the different noise schedules compared-- linear, sigmoid, power and cosine:
  \begin{align*}
  \textbf{Linear:}\quad
  k &= \frac{i}{K}, \\
  \textbf{Sigmoid:}\quad
  k &= \sigma\left(\alpha\left(\frac{i}{K} - 0.5\right)\right),\\
  \textbf{Power:}\quad
  k &= \left(\frac{i}{K}\right)^{0.5}, \\
  \textbf{Cosine:}\quad
  k &= 1 - \cos\left(\frac{i\pi}{2K}\right), \\
  \end{align*}
where $\sigma(\cdot)$ is the logistic sigmoid and $\alpha$ controls the sharpness.
For the sigmoid schedule, we normalize endpoints to be $t_0=0$ and $t_K=1$.
The linear schedule allocates steps uniformly, cosine and power allocate more budget near low-noise regions, and sigmoid concentrates updates around the middle of the trajectory.

\para{Rectified-Flow}
To further reduce inference time and NFE for downstream tasks like test-time planning, we used ReFlow~\cite{liu2022flow} to post-train \oursft model, and call it \ours-ReFlow.
The teacher and student model are initialised with a \oursft model where we freeze the teacher model. 
At each training iteration, we sample noise~$x^0$ and predict future latents with the teacher model~$\hat{x}^1$.
This student model is updated with the predicted latent as the target using mean squared error loss given by: $\mathcal{L}^{\texttt{ReFlow}}(\phi)=\mathbb{E}_{x^0_t, \hat{x}^1_t, \tau}\left[\left\|
    (\hat{x}_t^1 - x_t^0)
    -
    \dyn(\latenthist_t, \latentmem_t, \acttraj_t, x^{\tau}_t, \tau)
    \right\|_2^2\right]$.
The post-training with rectified flow is performed for 2K gradient steps on $4\times H100$~(6 hours) with a learning rate of $2e^{-5}$.

\section{Additional World Model Evaluation Results}
\label{app:add_results}
We provide additional results to evaluate \ours at coherent generations, impact of KV Cache on inference time, benefits of noise schedules, finetuning on task data, and post-training with rectified flow. 

\subsection{World Model Evaluation}\label{app:world_model_eval}
For a trajectory in validation dataset, we generated the rollout from the 20-th step, and compute the metrics using the generations for next 10s. We use the first 20 frames to initialize the memory and history for \ours and \ctrlworld. 
We report LPIPS~\cite{zhang2018perceptual}, FID~\cite{heusel2017gans}, and FVD~\cite{unterthiner2018towards} obtained using the ground truth videos. 
To obtain LPIPS, we utilize the functionality in \texttt{torchmetrics}\footnote{\texttt{https://github.com/Lightning-AI/torchmetrics}} that uses the per-frame features obtained from \texttt{vgg} layers.
To compute the FID, we use the implementation provided in \texttt{pytorch-fid}\footnote{\texttt{https://github.com/mseitzer/pytorch-fid}} repository.
Our results on FVD are computed using the \texttt{Style-GAN-V}\footnote{\texttt{https://github.com/universome/stylegan-v}} repository. Here, we subsample multiple trajectories of 16 frames with a stride of 8 from each trajectory.

\begin{table}[th]
    \centering
    \setlength{\tabcolsep}{3.2pt}
    \renewcommand{\arraystretch}{1.05}
    \footnotesize
    \begin{tabular}{l c ccc ccc c}
    \toprule
    & & \multicolumn{3}{c}{\textit{Exterior}} & \multicolumn{3}{c}{\textit{Wrist}} & \\
    \cmidrule(lr){3-5} \cmidrule(lr){6-8}
    \textbf{Method} & \textbf{NFE}
    & \textbf{LPIPS} $\downarrow$ & \textbf{FID} $\downarrow$ & \textbf{FVD} $\downarrow$
    & \textbf{LPIPS} $\downarrow$ & \textbf{FID} $\downarrow$ & \textbf{FVD} $\downarrow$
    & \textbf{Time (s)} $\downarrow$ \\
    \midrule
    \rowcolor{pastellavender}
    \multicolumn{9}{c}{\textbf{\textit{\textcolor{slategray}{DROID}}}} \\
    Ctrl-World & 8  & 0.169 & 31.63 & 116.14 & 0.407 & 52.40 & 347.69 & 8.14 \\
               & 16 & 0.165 & 26.09 & 78.73  & 0.392 & 33.83 & 195.37 & 14.65 \\
               & 32 & 0.168 & 23.63 & 63.55  & 0.389 & 27.14 & 114.87 & 27.67 \\
               & 50 & 0.168 & 22.44 & 55.05  & 0.388 & 25.32 & 91.77  & 42.33 \\
    \midrule
    \ours      & 8  & \cellcolor{lightblue}\textbf{0.117} & 10.59 & 28.97 & \cellcolor{lightblue}0.372 & 24.25 & 104.53 &
  \cellcolor{lightblue}\textbf{2.53} \\
               & 16 & \cellcolor{lightblue}\textbf{0.117} & 10.20 & 27.83 & \cellcolor{lightblue}\textbf{0.371} & 21.50 & 90.72 & 4.78 \\
               & 32 & \cellcolor{lightblue}0.120 &\cellcolor{lightblue} 9.67 & \cellcolor{lightblue}\textbf{25.94} & \cellcolor{lightblue}0.378 & \cellcolor{lightblue}17.53 & \cellcolor{lightblue}\textbf{63.36}
  & 9.22 \\
               & 50 & 0.122 & \cellcolor{lightblue}\textbf{9.51 }& \cellcolor{lightblue}26.54 & \cellcolor{lightblue}0.378 & \cellcolor{lightblue}\textbf{16.75} & 66.89
  & 14.25 \\

    \midrule
    \rowcolor{pastelmint}
    \multicolumn{9}{c}{\textbf{\textit{\textcolor{slategray}{New dataset}}}} \\
    Ctrl-World & 8  & 0.193 & 48.90 & 226.29 & 0.374 & 51.26 & 434.84 & 8.14 \\
               & 16 & 0.182 & 36.16 & 139.54 & 0.366 & 38.76 & 277.13 & 14.65 \\
               & 32 & 0.183 & 32.18 & 105.38 & 0.365 & 33.73 & 173.15 & 27.67 \\
               & 50 & 0.184 & 31.44 & 91.48  & 0.367 & 33.47 & 145.86 & 42.33 \\
    \midrule
    \ours      & 8  & \cellcolor{lightblue}\textbf{0.154} & \cellcolor{lightblue}23.89 & 89.55 & \cellcolor{lightblue}\textbf{0.364} & 31.70 &
  193.55 & \cellcolor{lightblue}\textbf{2.53} \\
   & 16 & \cellcolor{lightblue} 0.155 & \cellcolor{lightblue}23.95 & 88.27 & \cellcolor{lightblue}\textbf{0.364} & 30.77 & 184.62 & 4.78 \\
   & 32 & \cellcolor{lightblue}0.157 & \cellcolor{lightblue}\textbf{23.45} & 92.36 & \cellcolor{lightblue}0.365 & 28.24 & \cellcolor{lightblue}148.85 & 9.22 \\
   & 50 & \cellcolor{lightblue}0.159 & \cellcolor{lightblue}23.48 & \cellcolor{lightblue}\textbf{87.03} & \cellcolor{lightblue}0.371 & \cellcolor{lightblue}\textbf{27.37} &
  \cellcolor{lightblue}\textbf{145.04} & 14.25 \\
    \bottomrule
  \end{tabular}
  \caption{Comparison of \ours and \ctrlworld at LPIPS, FID and FVD metrics. \ours generates with higher fidelity than Ctrl-World and has significantly  better performance at low inference budgets. Here, NFE is Number of Function Evaluations and inference time is the time required to generate a 10s segment on a single H100 GPU.}
    \label{tab:results_full}
  \end{table}
  
\subsection{Quantitative Results}\label{app:quantitative_results}
\begin{figure}[h]
    \centering
    \includegraphics[width=\linewidth]{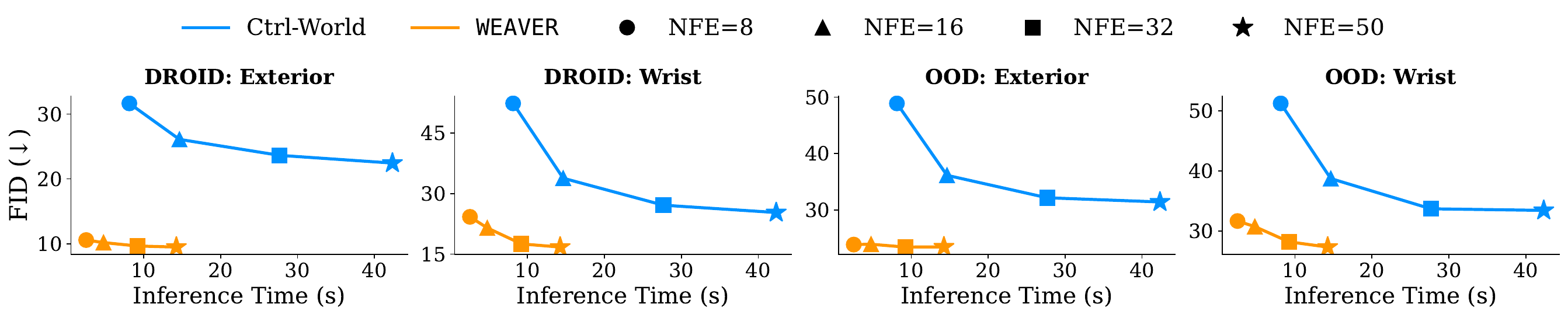}
    \caption{We compare FID vs inference time for \ours and \ctrlworld and find that \ours outperforms the baseline with upto $16\times$ more inference time.}
    \label{fig:fid_inference_time}
\end{figure}

\begin{figure}[h]
    \centering
    \includegraphics[width=\linewidth]{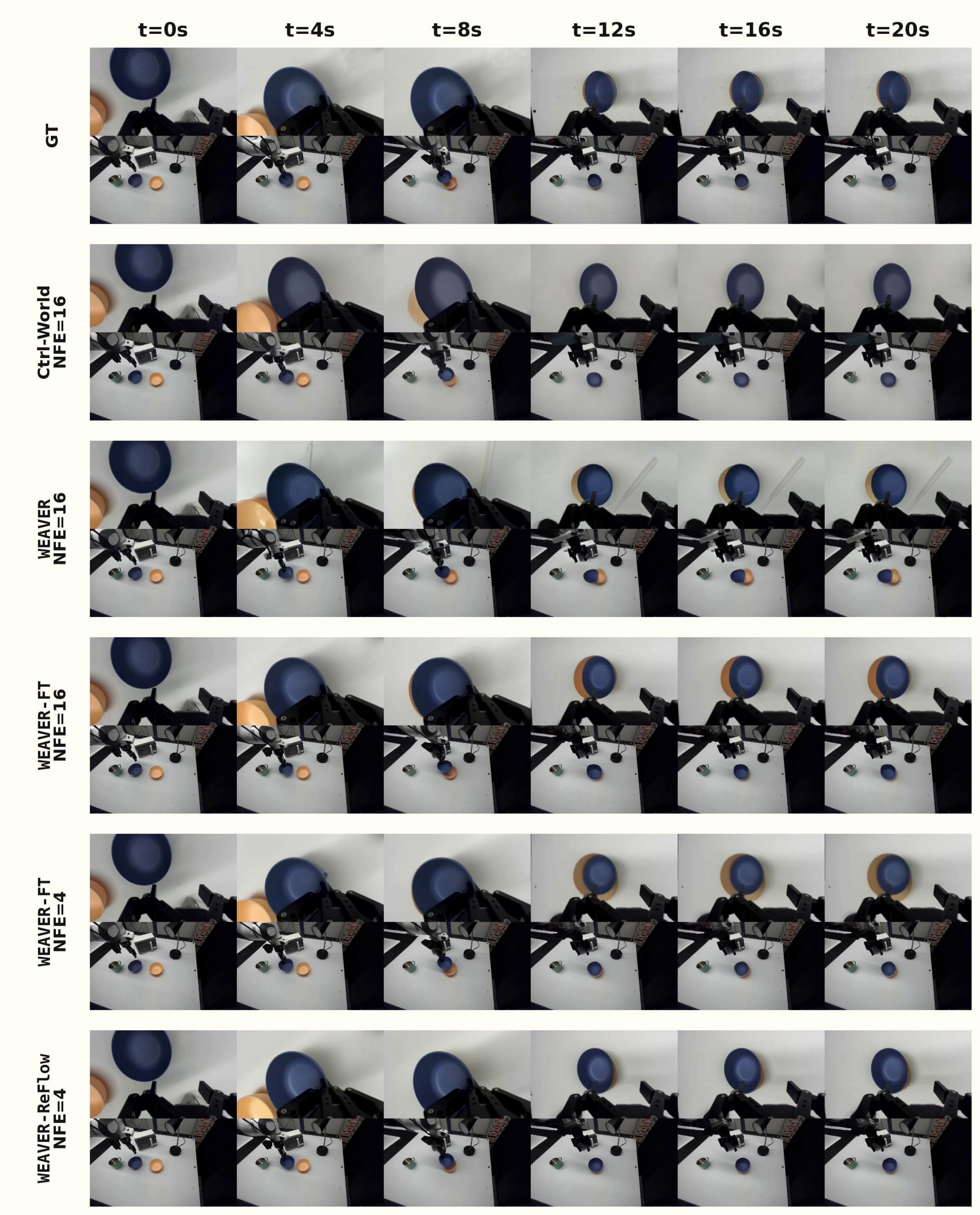}
    \caption{We compare the rollouts on task obtained from \ctrlworld, \ours, \oursft and \oursreflow at different NFE values of 4 and 16. We generate rollouts for 20 seconds and present predicted camera views at every 4 second. We observe that \oursreflow is better than \oursft at NFE=4 and has comparable performance with other models using NFE=16.}
    \label{fig:stack_all_app_nfe}
\end{figure}

\begin{figure}[h]
    \centering
    \includegraphics[width=\linewidth]{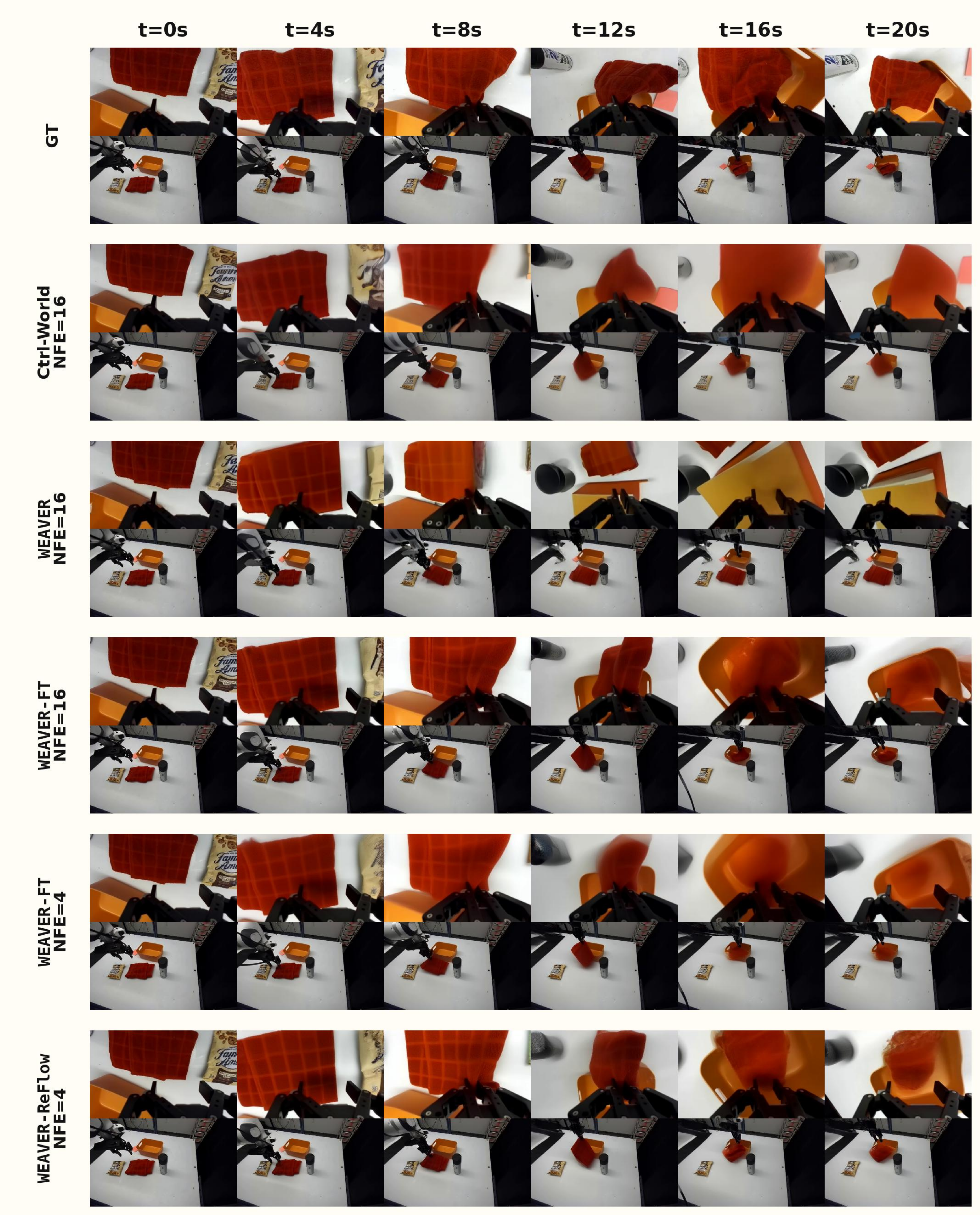}
    \caption{We compare the rollouts on task obtained from \ctrlworld, \ours, \oursft and \oursreflow at different NFE values of 4 and 16. We generate rollouts for 20 seconds and present predicted camera views at every 4 second. We observe that \oursreflow is better than \oursft at NFE=4 and is more consistent than \ctrlworld and \ours with NFE=16.}
    \label{fig:towel_all_app_nfe}
\end{figure}

Table~\ref{tab:results_full} reports the comparison of \ours and Ctrl-World on DROID and OOD datasets across multiple metrics. 
We observe that performance of Ctrl-World deteriorates with lower NFE whereas \ours shows slight drop in performance with decrease in NFE.
Moreover, with similar NFE values, our method is $3\times$ faster at generating rollouts than Ctrl-World.
In Fig.~\ref{fig:fid_inference_time}, we present the comparison of FID and inference time and observe that \ours with lowe NFE of 8 outperforms \ctrlworld with large NFE of 50. We also include qualitative results of different NFEs with different world models in Fig.~\ref{fig:stack_all_app_nfe} and Fig.~\ref{fig:towel_all_app_nfe}.

 \begin{table}[h]
    \centering
    \setlength{\tabcolsep}{3.2pt}
    \renewcommand{\arraystretch}{1.05}
\begin{tabular}{l|cccc}
    \toprule
    \textbf{ /\ NFE} & \textbf{8} & \textbf{16} & \textbf{32} & \textbf{50} \\
    \midrule
    Without KV cache & 3.30 & 6.54 & 12.98 & 20.21 \\
    With KV cache    & 2.53 & 4.78 & 9.22  & 14.25 \\
    \bottomrule
  \end{tabular}
    \caption{We report the inference time (in seconds) taken to generate a 10s trajectory on a single H100 GPU at different NFE, and observe that using KV Cache can reduce inference time by upto ~30\%. }
    \label{tab:inference_kvcache}
  \end{table}

\subsection{Reducing inference time with KV Cache}
\label{app:kv_cache_results}
During the iterative denoising process, the latents of memory and history frames are passed with a constant noise level $k=1$.
Since it does not change during this process, we compute the cache of key-value vectors of memory and history latents at the first denoising step.
In Table~\ref{tab:inference_kvcache}, we report that KV Cache can reduce inference time by up to 30\%. 

\begin{table}
    \centering
    \setlength{\tabcolsep}{3.2pt}
    \renewcommand{\arraystretch}{1.05}
    \footnotesize
    \begin{tabular}{l c ccc ccc}
      \toprule
      & & \multicolumn{3}{c}{\textit{Exterior}} & \multicolumn{3}{c}{\textit{Wrist}} \\
      \cmidrule(lr){3-5} \cmidrule(lr){6-8}
      \textbf{Method} & \textbf{Schedule}
      & \textbf{LPIPS} $\downarrow$ & \textbf{FID} $\downarrow$ & \textbf{FVD} $\downarrow$
      & \textbf{LPIPS} $\downarrow$ & \textbf{FID} $\downarrow$ & \textbf{FVD} $\downarrow$ \\
      \midrule
      \rowcolor{pastellavender}
      \multicolumn{8}{c}{\textbf{\textit{\textcolor{slategray}{DROID}}}} \\
      Ctrl-World & linear
        & 0.165 & 26.09 & 78.73
        & 0.392 & 33.83 & 195.37 \\
      \ours & linear
        & \cellcolor{lightblue}\textbf{0.117} & 11.32 & \cellcolor{lightblue}\textbf{26.38}
        & \cellcolor{lightblue}0.375 & 24.43 & 98.82 \\
      & sigmoid
        & \cellcolor{lightblue}\textbf{0.117} & \cellcolor{lightblue}10.88 & 29.76
        & \cellcolor{lightblue}0.375 & 22.69 & 104.89 \\
      & power
        & \cellcolor{lightblue}\textbf{0.117} & \cellcolor{lightblue}10.57 & 27.93
        & \cellcolor{lightblue}\textbf{0.369} & \cellcolor{lightblue}\textbf{21.17} & \cellcolor{lightblue}91.51 \\
      & cosine
        & \cellcolor{lightblue}\textbf{0.117} & \cellcolor{lightblue}\textbf{10.20} & 27.83
        & \cellcolor{lightblue}0.371 & \cellcolor{lightblue}21.50 & \cellcolor{lightblue}\textbf{90.72} \\

      \midrule
      \rowcolor{pastelmint}
      \multicolumn{8}{c}{\textbf{\textit{\textcolor{slategray}{New dataset}}}} \\
      Ctrl-World & linear
        & 0.182 & 36.16 & 139.54
        & \cellcolor{lightblue}0.366 & 38.76 & 277.13 \\
      \ours & linear
        & \cellcolor{lightblue}0.157 & 25.37 & 96.16
        & \cellcolor{lightblue}0.367 & 33.24 & 217.65 \\
      & sigmoid
        & \cellcolor{lightblue}0.156 & 24.83 & 93.15
        & \cellcolor{lightblue}0.367 & 32.48 & 216.89 \\
      & power
        & \cellcolor{lightblue}\textbf{0.155} & \cellcolor{lightblue}\textbf{23.82} & \cellcolor{lightblue}\textbf{84.91}
        & \cellcolor{lightblue}\textbf{0.363} & \cellcolor{lightblue}31.60 & \cellcolor{lightblue}185.30 \\
      & cosine
        & \cellcolor{lightblue}\textbf{0.155} & \cellcolor{lightblue}23.95 & 88.27
        & \cellcolor{lightblue}0.364 & \cellcolor{lightblue}\textbf{30.77} & \cellcolor{lightblue}\textbf{184.62} \\
      \bottomrule
    \end{tabular}
    
    \caption{ We see that cosine and power noise schedules allocates higher budget at low noise scales and perform better than linear and sigmoid schedules. The numbers are reported with NFE=16.}
    \label{tab:noise_schedule}
  \end{table}

\subsection{Noise schedules during inference}\label{app:noise_schedule_result}Table~\ref{tab:noise_schedule} compares different noise schedules where we observe that both power and cosine noise schedules perform better than sigmoid and linear noise schedules. 
Since the world model needs to generate with higher fidelity, the noise schedules that allocate more bandwidth at low noise regions aids in generating fine-grained details. 

 \begin{table}
    \centering
    \setlength{\tabcolsep}{3.2pt}
    \renewcommand{\arraystretch}{1.05}
    \footnotesize
    \begin{tabular}{l c ccc ccc}
      \toprule
      & & \multicolumn{3}{c}{\textit{Exterior}} & \multicolumn{3}{c}{\textit{Wrist}} \\
      
      \cmidrule(lr){3-5} \cmidrule(lr){6-8}
      \textbf{Method} & \textbf{NFE}
      & \textbf{LPIPS} $\downarrow$ & \textbf{FID} $\downarrow$ & \textbf{FVD} $\downarrow$
      & \textbf{LPIPS} $\downarrow$ & \textbf{FID} $\downarrow$ & \textbf{FVD} $\downarrow$ \\
      \midrule
      \rowcolor{pastelmint}
      \multicolumn{8}{c}{\textbf{\textit{\textcolor{slategray}{Task Data (OOD)}}}} \\
      \ctrlworld & 50
        & 0.184 & 31.44 & 91.48
        & 0.367 & 33.47 & 145.86 \\
      \ctrlworldft & 16
        & 0.140 & 29.89 & 81.16
        & 0.295 & 34.77 & 283.00 \\
      & 50
        & 0.142 & 25.96 & 58.27
        & 0.292 & 25.80 & 134.69 \\
      \midrule
      \ours & 50
        & 0.159 & 23.48 & 87.03
        & 0.371 & 27.37 & 145.04 \\
      \oursft & 4
        & 0.116 & 16.62 & 50.68
        & 0.304 & 33.32 & 219.15 \\
      & 16
        & 0.118 & 14.24 & 40.50
        & 0.303 & 23.39 & 146.35 \\
       & 50
        & 0.121 & 13.69 & 40.09
        & 0.308 & 18.73 & 99.81 \\
    \midrule
       \oursreflow & 4
        & 0.123 & 14.95 & 44.30
        & 0.312 & 23.56 & 138.67 \\
      \bottomrule
    \end{tabular}
    \caption{Compare the finetuned variants of \ours and \ctrlworld on Task dataset (OOD) where \oursft outperforms the baselines. Moreover, the post-training step (\oursreflow) further helps to reduce inference budget.}
    \label{tab:finetuning}
  \end{table}

\subsection{Finetuning}\label{app:finetuning_result}
\begin{figure}[h]
    \centering
    \includegraphics[width=\linewidth]{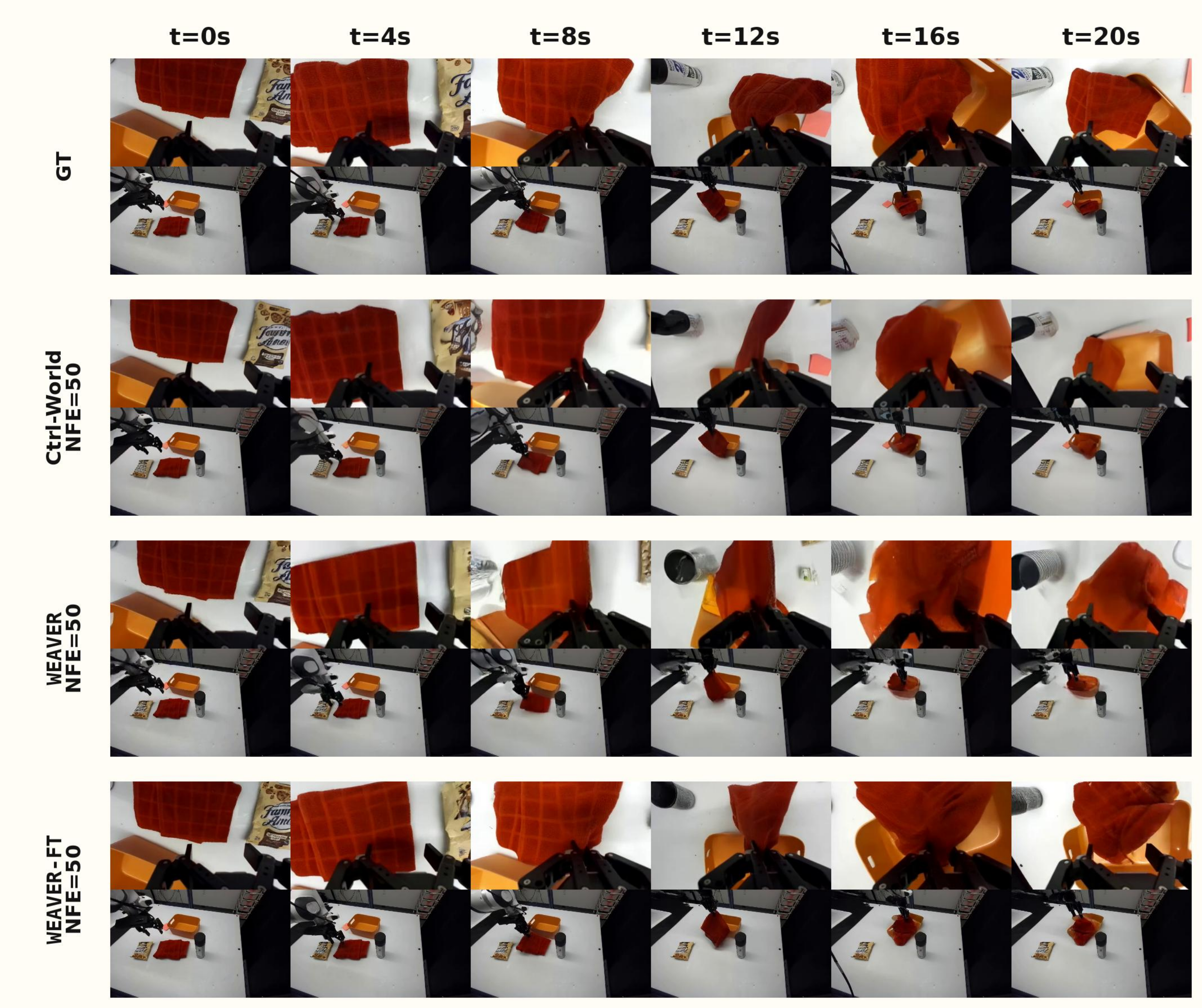}
    \caption{We compare the rollouts on task obtained from \ctrlworld, \ours and \oursft and \oursreflow at different NFE=50. We generate rollouts for 20 seconds and present predicted camera views at every 4 second. We observe that \ctrlworld struggles at retraining information about the towel after 12 seconds and \oursft is more consistent with the ground truth.}
    \label{fig:towel_all_app}
\end{figure}

\begin{figure}[h]
    \centering
    \includegraphics[width=\linewidth]{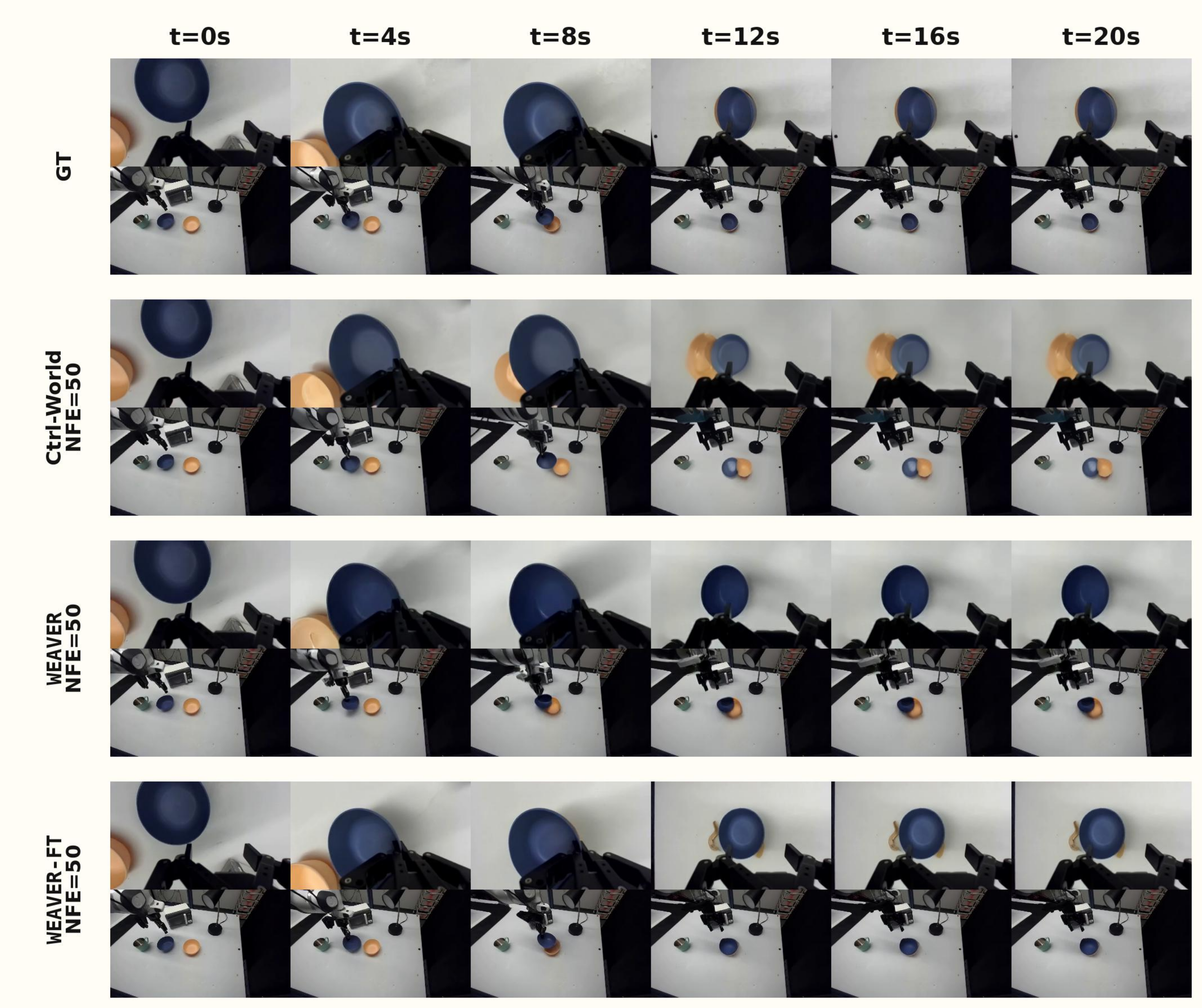}
    \caption{We compare the rollouts on task obtained from \ctrlworld, \ours and \oursft and \oursreflow at different NFE=50. We generate rollouts for 20 seconds and present predicted camera views at every 4 second. We observe that \ours and \oursft is better than \ctrlworld at predictions.}
    \label{fig:stack_all_app}
\end{figure}

\begin{figure}[h]
    \centering
    \includegraphics[width=\linewidth]{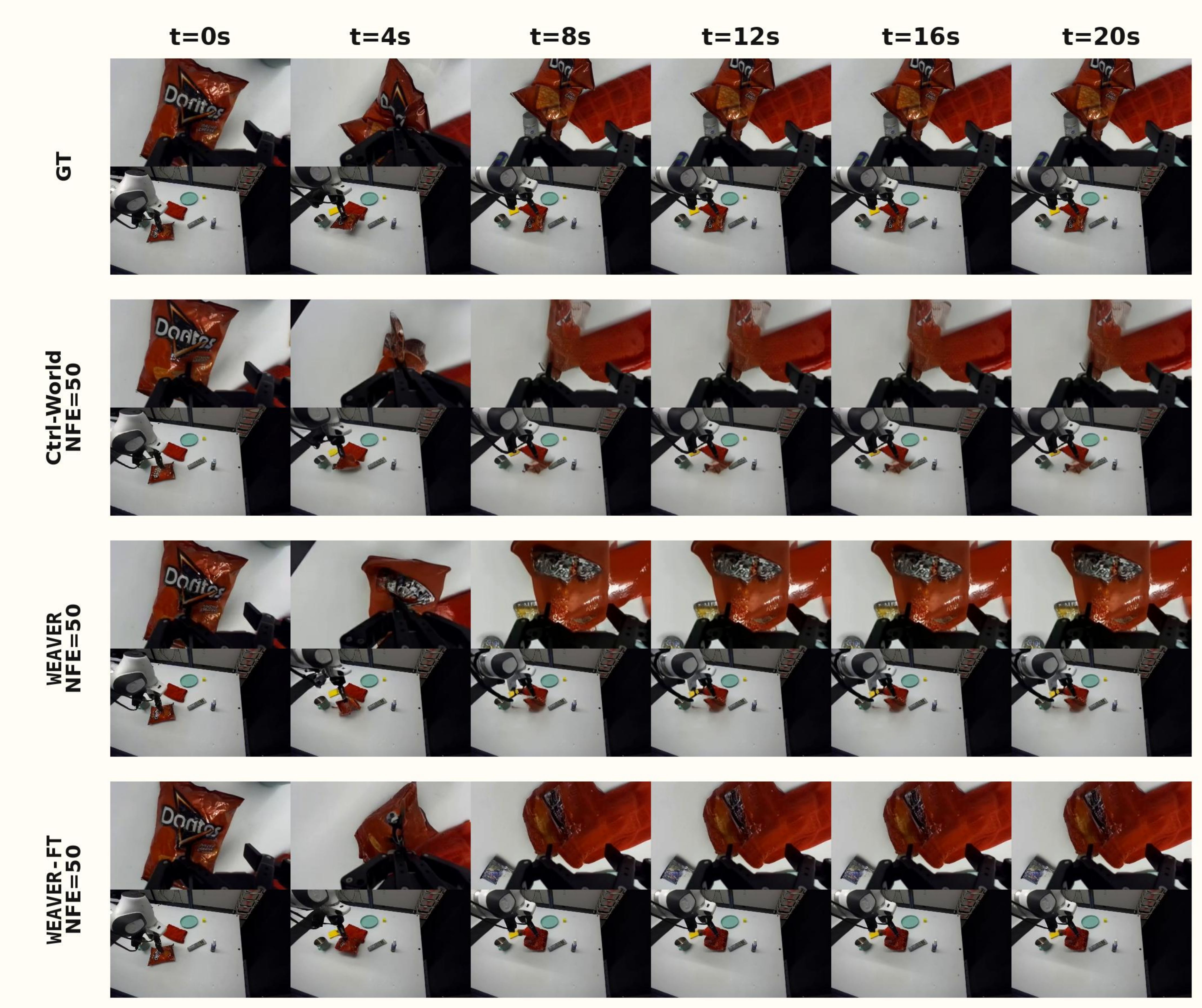}
    \caption{We compare the rollouts on task obtained from \ctrlworld, \ours and \oursft and \oursreflow at different NFE=50. We generate rollouts for 20 seconds and present predicted camera views at every 4 second. We see that \ctrlworld struggles to predict the object from t=4s compared to \ours and \oursft.}
    \label{fig:bag_all_app}
\end{figure}

In Table~\ref{tab:finetuning}, we observe that finetuning significantly improves performance and with NFE=16 it performs better than \ours with NFE=50. 
To provide a fair comparison with Ctrl-World, we finetune the baseline for 20K gradient steps on $4\times H100$ and see that the finetuned \ctrlworld (called \ctrlworldft) performs better than the pretrained model. 
However, \ours-FT ourperforms Ctrl-World-FT across metrics and the performance is still larger with low NFE=16.
This further demonstrates that finetuning does not help in reducing inference time for Ctrl-World. We also provide qualitative results of the rollouts generated from \ctrlworld, \ours, \oursft in Fig.~\ref{fig:towel_all_app}, Fig.~\ref{fig:stack_all_app} and Fig.~\ref{fig:bag_all_app}.

\subsection{Posttraining with Rectified Flow}\label{app:rectified_flow_result}
In table~\ref{tab:finetuning}, we present the results of \ours-ReFlow with small inference budget and observe that it reduces the performance gap with \oursft evaluated with a large NFE=16. 
This makes it suitable for test-time steering as observed in Section~\ref{sec:faster_inference}.

\begin{table}[t]
\centering
\small
\caption{Inference time breakdown for test-time planning on A6000 Ada GPU. We report the runtime of each component with different horizons of world model imaginations across 10 function calls.}
\label{tab:inference_time}
\begin{tabular}{lcccc}
\toprule
\textbf{Component} & \textbf{Notation} & \textbf{Batch Size} & \textbf{Horizon} & \textbf{Runtime (s) $\downarrow$}  \\
\midrule
Policy sampling & $\policy_\theta(\action_{t:t+h}\mid \obs_t,\lang)$ & &-- & $0.1979_{\pm 0.0002}$ \\
\hline
Dynamics model  & $\dyn(\hat{\latent}_{t+1:t+h}\mid \latentmem_t, \latenthist_t, \action_{t:t+h})$ & 4 & 9 & $1.0203_{\pm 0.0100}$ \\
(\ours)& & 4 &  12 &$1.2493_{\pm 0.0170}$ \\
&&  4 &  15 & $1.4547_{\pm 0.0160}$ \\
&&  1 &  15 & $0.4476_{\pm 0.0049}$ \\ 
\hline
Dynamics Model  & $\dyn(\hat{\latent}_{t+1:t+h}\mid \latentmem_t, \latenthist_t, \action_{t:t+h})$&  4 &  15 & $29.4244_{\pm 1.0162}$\\
(\ctrlworld) &&  1 &  15 & $7.4236_{\pm 0.1201}$ \\
\hline
Reward inference & $\reward(\hat{\latent}_{t+1:t+h},\lang)$ &  4 &--& $0.0006_{\pm 0.0002}$  \\
\hline
Critic inference & $\critic(\hat{\latent}_{t+h},\lang)$ &  4  &--& $0.0005_{\pm 0.0001}$\\
\bottomrule
\end{tabular}
\end{table}

\section{Additional Downstream Application Results}
\subsection{Policy Evaluation Results}
\label{app:policy_evaluation}

\begin{wrapfigure}{r}{0.6\textwidth}
    \centering
    \vspace{-0.8cm}
    \includegraphics[width=\linewidth]{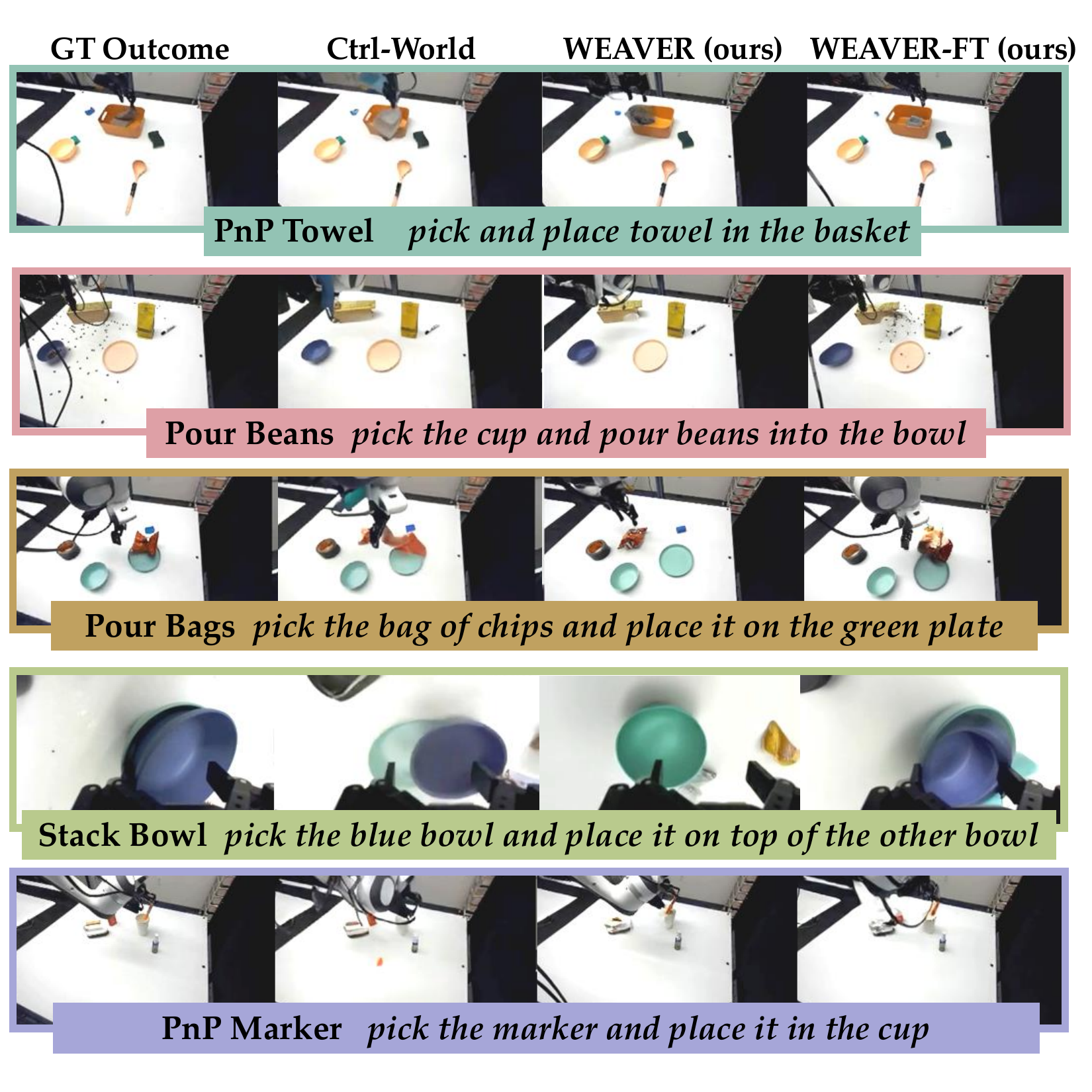}
    \caption{\textbf{Policy Evaluation Results}. We show policy evaluation results for all five tasks across three world models.}
    \label{fig:policy_eval_full}
    \vspace{-0.7cm}
\end{wrapfigure}

\begin{figure}
    \centering
    \includegraphics[width=\linewidth]{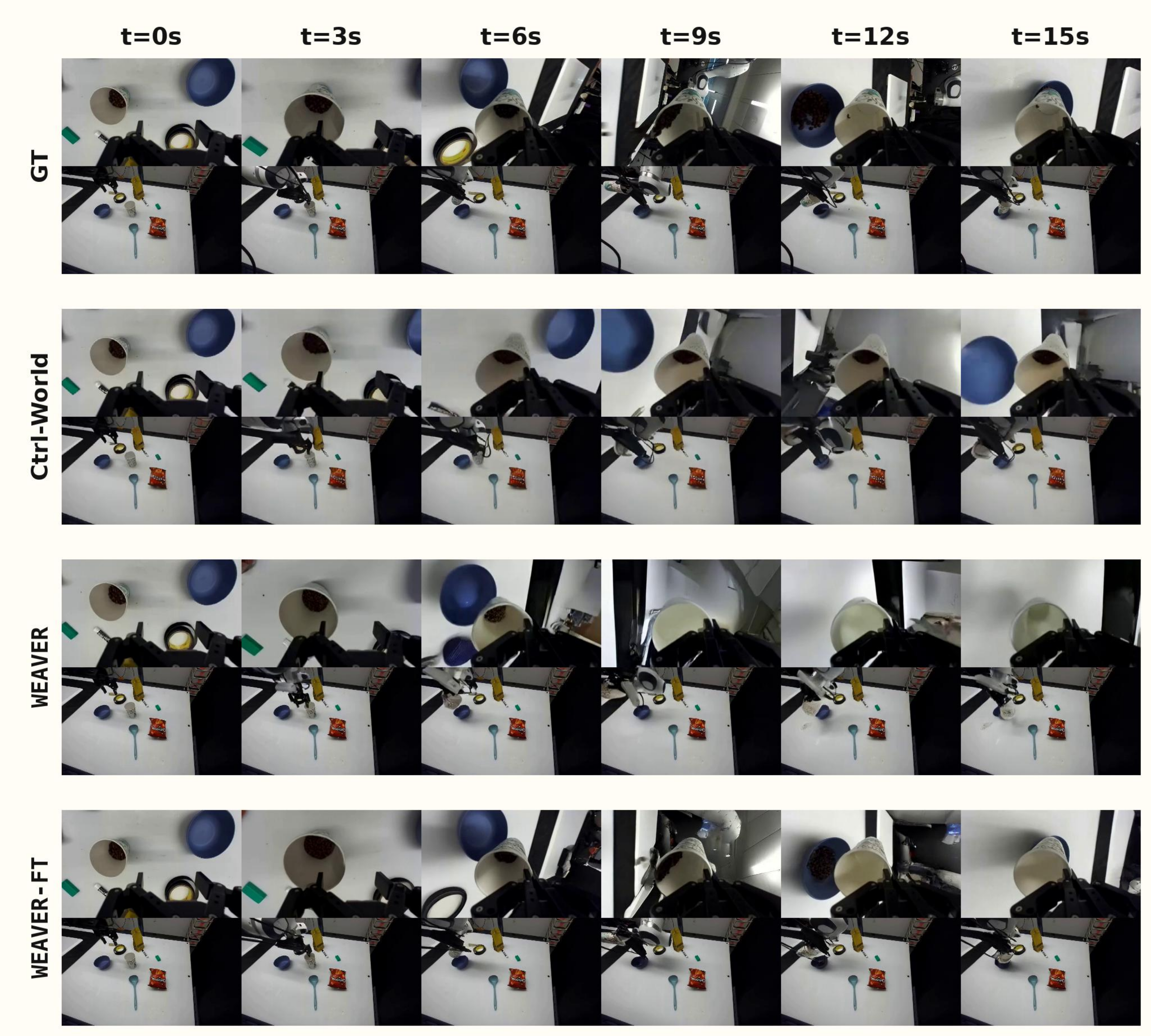}
    \caption{We compare the rollouts on Pour Beans task obtained from \ctrlworld, \ours and \oursft for policy evaluation at NFE=50. We generate rollouts for 15 seconds and present predicted camera views at every 3 second. We see that \ctrlworld and  \ours struggles to predict the beans in the bowl at t=12s compared to \oursft.}
    \label{fig:policy_qualitative_pour1}
\end{figure}

\begin{figure}
    \centering
    \includegraphics[width=\linewidth]{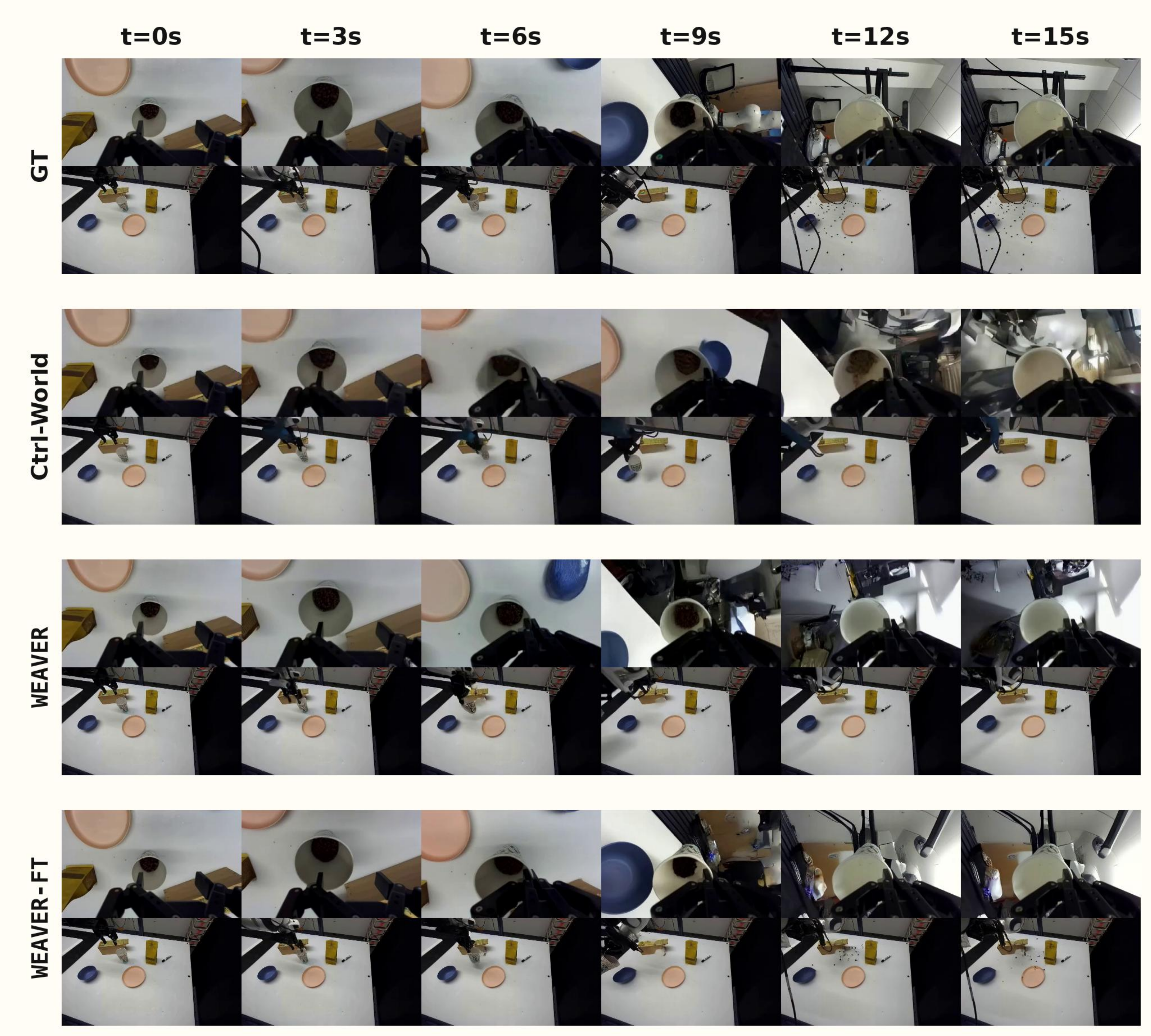}
    \caption{We compare the rollouts on Pour Beans task obtained from \ctrlworld, \ours and \oursft for policy evaluation at NFE=50. We generate rollouts for 15 seconds and present predicted camera views at every 3 second. We see that \ctrlworld and  \ours struggles to predict the beans on the table at t=15s compared to \oursft.}
    \label{fig:policy_qualitative_pour2}
\end{figure}

We provide the full policy evaluation rollouts in Fig.~\ref{fig:policy_eval_full}. Both \ctrlworld and \ours struggle to accurately predict policy performance, especially on challenging tasks involving dynamic manipulation, such as pouring beans, and deformable object manipulation, such as bag and towel manipulation. For the \textbf{PnP Bag} task, grasping the bag is particularly challenging because the world model must accurately infer the gripper depth across two camera views while also modeling the contact dynamics between the gripper and the deformable object. These challenges become more pronounced as the prediction horizon increases. In contrast, \oursft substantially improves evaluation accuracy through finetuning as shown in Fig.~\ref{fig:policy_qualitative_pour1} and Fig.~\ref{fig:policy_qualitative_pour2}. Future work could further improve long-horizon prediction by designing better memory and history representations, enabling the model to better reason about occlusions and deformable object dynamics.

In addition to Pearson correlation and MMRV, we also report RMSE and Spearman rank correlation~\citep{wissler1905spearman}. Across these metrics, we observe a consistent trend: \oursft achieves the strongest correlation and lowest prediction error. In addition, \ours outperforms \ctrlworld in zero-shot policy evaluation on out-of-distribution task dataset.
The full quantitative results are shown in Table~\ref{tab:reward_prediction_metrics}.

\begin{table}[t]
\centering
\caption{Comparison of reward prediction quality across different world models. We report RMSE, Spearman correlation, Pearson correlation, and MMRV.}
\label{tab:reward_prediction_metrics}
\begin{tabular}{lcccc}
\toprule
\textbf{Method} & \textbf{RMSE} $\downarrow$ & \textbf{Spearman} $\uparrow$ & \textbf{Pearson} $\uparrow$ & \textbf{MMRV} $\downarrow$ \\
\midrule
CtrlWorld & 0.410 & 0.523 & 0.552 & 0.215 \\
Ours      & 0.359 & 0.594 & 0.563 & 0.155 \\
Ours-FT   & \textbf{0.188} & \textbf{0.870} & \textbf{0.863} & \textbf{0.035} \\
\bottomrule
\end{tabular}
\end{table}

\subsection{Policy Improvement Results}
\label{app:policy_improvement}

\begin{figure}
    \centering
    \includegraphics[width=\linewidth]{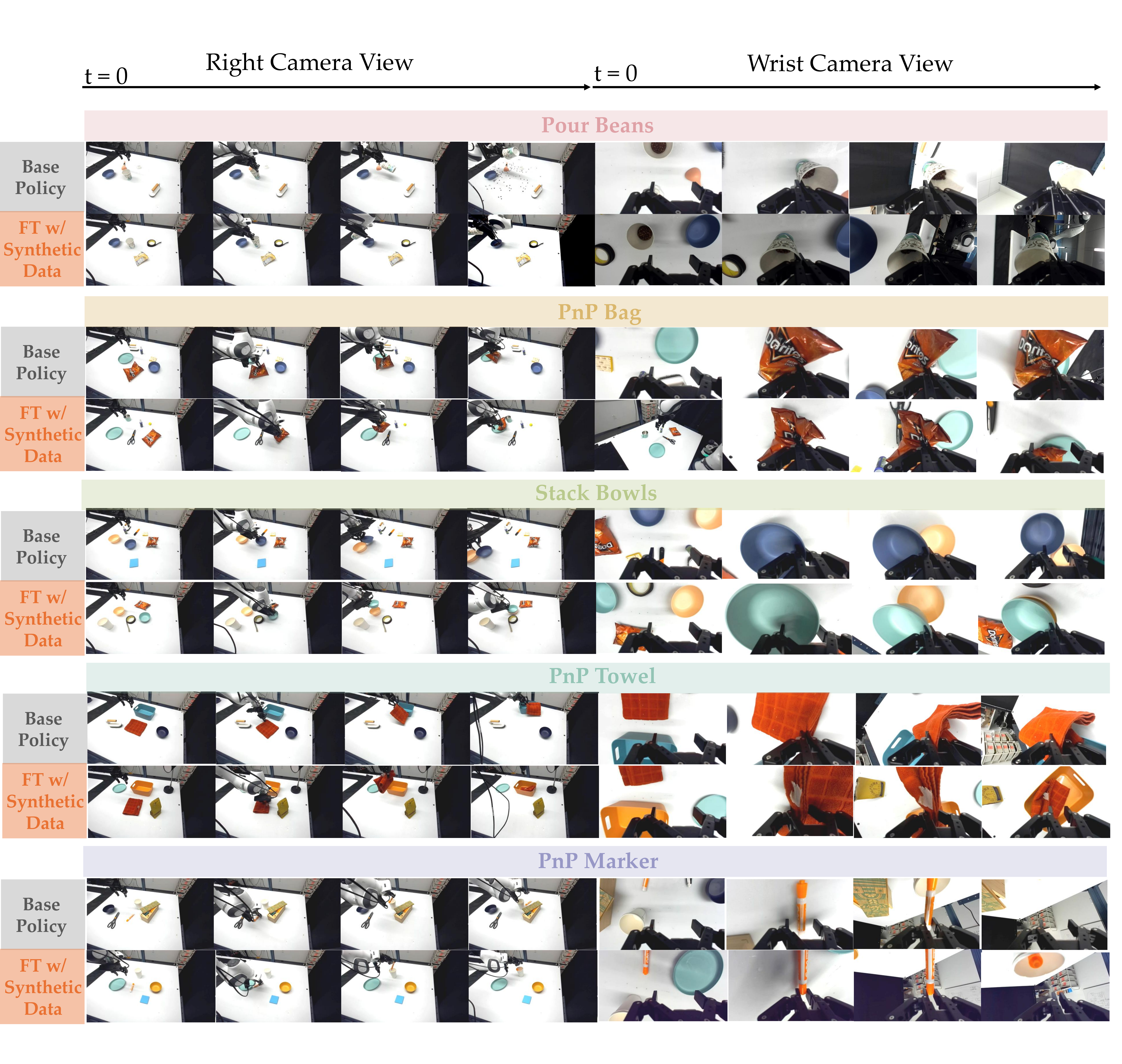}
    \caption{\textbf{Policy Improvement Results}. We demonstrate the rollouts for five tasks among the base policy and policy FT w/ Synthetic Data. With \ours generated synthetic data, policy finetuning can improvement upon all tasks.}
    \label{fig:policy_improvement_full}
\end{figure}

We provide additional qualitative results of policy improvement in Fig.~\ref{fig:policy_improvement_full}. These examples show that the base policy often suffers from imprecise grasping and placement, as well as insufficient adjustment during dynamic manipulation. We also observe that the base policy tends to produce larger per-step motions, resulting in unstable robot control. In contrast, the finetuned policy substantially reduces these large movements and sharpens the action distribution, leading to smoother and more stable execution.

We also note that the RoboMeter reward labels are not perfect. For the \textbf{PnP Marker} task, we observe cases where the reward model fails to distinguish fine-grained placement accuracy, which can introduce noise into the predicted rewards. Future work could improve reward supervision by collecting more diverse failure data to train a more general and precise reward model. To mitigate the effect of noisy reward labels, we set the advantage threshold to $0.1$, which helps prevent low-quality segments from being selected for finetuning and potentially degrading policy performance. As shown in Fig.~\ref{fig:reward_plot}, our filtering procedure is able to select the best action samples among the candidates.

\subsection{Test-Time Planning Results}
\label{app:inference-time-latency}
\para{Inference-Time Latency}
Table~\ref{tab:inference_time} reports the inference-time breakdown of test-time planning on an A6000 Ada GPU. Overall, the runtime is dominated by the dynamics model imagination, while reward and critic inference are negligible, taking less than $0.001$s each. For \ours, the dynamics runtime increases moderately with the imagination horizon: from $1.0203$s at horizon $9$, to $1.2493$s at horizon $12$, and $1.4547$s at horizon $15$ with batch size $4$. Including policy sampling, reward inference, and critic inference, the full planning latency is approximately $1.22$s, $1.45$s, and $1.65$s for horizons $9$, $12$, and $15$, respectively.

Compared to \ctrlworld, \ours substantially reduces latency during imagination. At horizon $15$ and batch size $4$, \ours takes $1.4547$s for dynamics prediction, while \ctrlworld requires $29.4244$s, corresponding to a $20.2\times$ speedup. The same trend holds at batch size $1$, where \ours takes $0.4476$s compared to $7.4236$s for \ctrlworld, yielding a $16.6\times$ speedup. These results show that \ours enables substantially lower-latency test-time planning, making repeated world-model imagination practical during policy execution.

\section{Limitations}
\label{app:limitations}
While \ours demonstrates the promise of large-scale world models for policy evaluation, policy improvement, and test-time planning, several limitations remain.
\subsection{Partial Observability}Our world model relies primarily on visual observations, which provide only partial access to the underlying physical state. During manipulation, task-relevant information such as object contacts, grasp stability, applied forces, or occluded object geometry may be hidden from all available camera views. This limitation is especially pronounced for wrist-camera observations, where the viewpoint changes continuously, and for cluttered scenes where objects may leave the field of view or become occluded by the gripper. Although memory and multi-view conditioning mitigate this issue, purely visual prediction may still fail when the missing state cannot be inferred from image history alone. Incorporating additional sensing modalities, such as tactile feedback, force-torque sensing, or depth, may improve state estimation and long-horizon prediction under occlusion.
\subsection{Complex Deformable and Dynamic Interactions}Deformable-object manipulation and dynamic manipulation remain challenging for learned world models. Objects such as towels, bags, and granular materials exhibit high-dimensional, history-dependent dynamics that are difficult to capture from limited robot data. Small errors in predicted contact, grasp location, or object configuration can compound over time and lead to qualitatively incorrect rollouts. This is particularly evident in tasks such as pouring, where the motion of granular material depends sensitively on cup pose, velocity, and contact with the container. Future work may improve prediction fidelity by incorporating physics priors, hybrid neural-physics models, or neural simulators specialized for deformable and granular dynamics.
\subsection{Limited Planning Horizon at Test Time}Although our inference acceleration strategies make test-time planning feasible with a large generative world model, latency still limits online planning to a single action chunk. As a result, the planner can improve near-term action selection but cannot yet perform long-horizon lookahead. This restricts its ability to reason about delayed consequences or multi-stage recovery behaviors. Further improvements in sampling efficiency, model distillation, value estimation, or hierarchical planning could enable longer-horizon online reasoning while maintaining real-time control.

\subsection{Data Coverage and Embodiment Diversity}Our world model is pretrained primarily on DROID, which provides large-scale robot interaction data but is still tied to a specific robot embodiment and data collection setup. This may limit generalization to substantially different robots, camera configurations, end-effectors. In addition, some task dynamics in our evaluation, such as granular pouring, are underrepresented in the pretraining data. Scaling world-model training to more diverse sources, including cross-embodiment robot datasets, simulation data, and human videos, may improve robustness and broaden the range of behaviors that can be accurately imagined.

\subsection{Noisy Reward Supervision}Our latent reward and critic heads are trained using labels from an off-the-shelf reward model. While this enables efficient latent-space evaluation, the resulting supervision can be noisy or incomplete, especially for subtle failure modes. For example, a reward model may fail to distinguish between visually similar but semantically different outcomes, or may be insensitive to small errors in contact, placement, or task completion. Such noise can affect both policy evaluation and downstream policy improvement. A more reliable reward model trained on large-scale robot success and failure data, potentially with calibrated uncertainty, would likely improve the reliability of imagined rollout evaluation.

Overall, these limitations suggest that future progress will require not only larger and faster world models, but also richer sensing, broader data coverage, stronger physical inductive biases, and more accurate reward supervision.
\clearpage


\end{document}